\documentclass[lettersize,journal]{IEEEtran}
\usepackage{algorithmic}
\usepackage{array}
\usepackage[caption=false,font=normalsize,labelfont=sf,textfont=sf]{subfig}
\usepackage{textcomp}
\usepackage{stfloats}
\usepackage{url}
\usepackage[utf8]{inputenc}
\usepackage{verbatim}
\usepackage{graphicx}
\usepackage{textcomp}
\usepackage{amsmath,amssymb,amsfonts}
\usepackage[backend=biber,style=ieee,natbib=true, minnames=1]{biblatex} 
\usepackage{mathtools}
\usepackage{booktabs}
\usepackage[ruled,vlined]{algorithm2e}
\usepackage{rotating}
\usepackage{multirow}
\usepackage{makecell}
\usepackage{dsfont}
\usepackage{pdfpages}
\usepackage[export]{adjustbox}
\usepackage{color, colortbl}
\DeclareMathOperator*{\argmax}{arg\,max}

\usepackage{balance}

\begin{document}

\title{Self-Supervised Time-Series Anomaly Detection Using Learnable Data Augmentation}

\author{Kukjin Choi,
        Jihun Yi,
        Jisoo Mok,
        and~Sungroh Yoon,~\IEEEmembership{Senior Member,~IEEE}
\thanks{
        K. Choi is with the Innovation Center, Samsung Electronics, Hwaseong, 18448, Korea, and also with the Department of Electrical and Computer Engineering, Seoul National University, Seoul, 08826, Korea (e-mail: kj21.choi@snu.ac.kr). 
        
		J. Yi, J. Mok, and S. Yoon (\textit{corresponding author}) are with the Department of Electrical and Computer Engineering, Seoul National University, Seoul, 08826, Korea (e-mail: t080205@snu.ac.kr; magicshop1118@snu.ac.kr; sryoon@snu.ac.kr).}
}%

\markboth{IEEE Transactions on Emerging Topics in Computational Intelligence}%
{Choi \MakeLowercase{\textit{et al.}}: Self-Supervised Time-Series Anomaly Detection Using Learnable Data Augmentation}


\maketitle

\begin{abstract}
Continuous efforts are being made to advance anomaly detection in various manufacturing processes to increase the productivity and safety of industrial sites.
Deep learning replaced rule-based methods and recently emerged as a promising method for anomaly detection in diverse industries.
However, in the real world, the scarcity of abnormal data and difficulties in obtaining labeled data create limitations in the training of detection models.
In this study, we addressed these shortcomings by proposing a learnable data augmentation-based time-series anomaly detection (LATAD) technique that is trained in a self-supervised manner.
LATAD extracts discriminative features from time-series data through contrastive learning.
At the same time, learnable data augmentation produces challenging negative samples to enhance learning efficiency.
We measured anomaly scores of the proposed technique based on latent feature similarities.
As per the results, LATAD exhibited comparable or improved performance to the state-of-the-art anomaly detection assessments on several benchmark datasets and provided a gradient-based diagnosis technique to help identify root causes.
\end{abstract}


\section{Introduction}
\label{sec:introduction}
\IEEEPARstart{F}{ault} detection and diagnosis are crucial for the productivity, quality, and safety of any industry.
Increase in system complexity and number of operations amplifies the probability of quality degradation and accidents.
Therefore, timely detection of faults and anomalies using principal indicators from the entire system is crucial for preventing potential accidents and economic losses.
For decades, manufacturing sites have actively collected data from sensors and leveraged them to improve their products, processes, and services.
However, the outdated methods lack accuracy and flexibility as they rely on basic statistics or the domain knowledge of an expert.

Recently, data-driven approaches that employ deep neural networks have been applied to various manufacturing fields as they effectively learn the dynamics of complex systems and successfully resolve multiple drawbacks.
Furthermore, they outperform conventional methods such as statistical- and machine-learning-based methods in diagnosing the state of facilities and systems, and analyze temporal data generated from numerous sensors.
Representative examples such as recurrent neural networks (RNN) and its variants, including long short-term memory (LSTM) and gated recurrent unit (GRU), have become the standard method for time-series data anomaly detection in the industry.
Convolutional neural networks, which are mainly used in computer vision tasks, are being applied as well.
In the early stages, these methods learned patterns in segmented time series as like images. 
However, recently, it has become possible to model long-term temporal contexts using temporal convolutional networks (TCN)~\cite{bai2018empirical} that adopt dilated and causal convolutions.
Ever since~\citet{bahdanau2014neural} proposed an attention mechanism, it has been actively used in almost all domains, including machine translation~\cite{vaswani2017attention}, computer vision~\cite{woo2018cbam}, and time series analysis~\cite{qin2017dual}.
These attention-based models can capture long-range dependence in parallel by paying attention to input weights that contribute more to the output, thereby overcoming the limitation where the RNN only references context in an ordered fashion.

Despite recent progresses, the real-world application of deep-learning models has been limited by unavailability of training data.
Conventionally, there are three problems related to training data.
First, collecting a large amount of labeled data is time- and resource-intensive.
Second, the obtained labels are sometimes biased toward the inspector.
Third, failure modes are extremely rare in tightly controlled modern manufacturing processes, and the resultant class imbalance between the normal and abnormal data hampers model training.

Many research topics have emerged and are being actively studied to address these shortcomings.
To illustrate, conventionally, an anomaly detection task employs an unsupervised learning strategy that does not use labels.
This uses normal class data for training and classifies abnormal data according to the discrepancy of the expected model output for the given input data in the inference time.

Recent unsupervised learning methods have explored three approaches for detecting anomalies in time-series data: reconstruction, forecasting, and similarity~\cite{choi2021deep}.
Reconstruction-based methods consist of an encoder and decoder, the encoder maps the original data onto a low-dimensional latent feature space, also known as latent feature space, whereas the decoder attempts to reconstruct its representations from projected low-dimensional space to the original input space.
This family of methods models underlying attributes of normal data by minimizing some form of reconstruction error.
However, its ultimate objective is to learn general information for reconstruction rather than classification. 
As a result, the learned representations may not be suitable for anomaly detection~\cite{pang2021deep}.
In addition, several studies~\cite{zong2018deep,zaheer2020old} have shown that autoencoders can reconstruct anomalies as well.
The forecasting-based method assumes that each element in the sequence contextually depends on the previous one.
Therefore, the prediction for the next step is conditioned on the previous temporal instances, and detecting anomalies are by measuring their deviation from the observation.
As a result, this method is sometimes vulnerable to long-term anomalies due to its step-wise prediction technique.
The similarity-based method measures how far the value derived by the model exists from the distribution or cluster of accumulated data.
It primarily performs feature-to-feature comparisons in low-dimensions since directly comparing input data requires a lot of computational costs; however, feature-to-feature comparisons in low dimensions are difficult to interpret intuitively.

In this study, we propose a novel self-supervised time-series anomaly detection framework called learnable data augmentation-based time-series anomaly detection (LATAD) to address the abovementioned data limitations and mitigate drawbacks of previous unsupervised learning methods.
To the best of our knowledge, this is the first study to adopt the self-supervised learning (SSL) method for time-series anomaly detection.
SSL is a relatively new technique of training a deep learning model with labels inherently obtained from the data itself.
Conventionally, in SSL, diverse transformations for given data augment positive examples, whereas negative examples are sampled from different classes.
The proposed model learns a high-level discriminative representation of data in a low dimensions by maximizing the mutual information shared by the input data and the positive examples, and simultaneously minimizing the mutual information with the negative ones.
In particular, it jointly considers the correlations of different univariate time series and the temporal dependencies within each time series to extract feature representations.
In addition, we utilized triplet margin loss for SSL in the latent feature space; it pulls a feature at each timestamp closer to the positive samples from the temporal neighborhood while pushing them further from the negative samples augmented with learnable neural networks.
The contributions are summarized as follows.
\begin{itemize}
    \item We propose a novel feature extractor with learnable sample generators in an SSL framework for time-series anomaly detection that derived discriminative feature representations under limited data conditions.
    \item We adopt self-attention on time and feature axes, to fuse inter-correlations among univariate time series with temporal context and capture the underlying attributes of the multivariate time series data.
    \item We perform empirical studies on multiple public benchmark datasets and proved that the proposed model effectively detects anomalies and has generalization capabilities that are compared to the recently proposed methods.
    \item We provide a gradient-based interpretation that describes the decision of the model and helps diagnose anomalies.
\end{itemize}

\section{Related works}
\label{sec:related}
Time-series anomaly detection is a complex task that has been studied for several decades.
Recently, research related to unsupervised deep learning methods for anomaly detection in multivariate time series has received much attention.

The deep autoencoding Gaussian mixture model (DAGMM)~\cite{zong2018deep} employs an autoencoder for yielding a representation vector and feeding it to the Gaussian mixture model. 
It uses the estimated sample energy as a reconstruction error; a high energy indicates high abnormality.
The multi-scale convolutional recursive encoder decoder (MSCRED)~\cite{zhang2019deep} jointly considers the time dependencies and the interpretation of anomaly severity.
It comprises a convolutional LSTM with an attention mechanism that reconstruct the input matrices from the feature map of each layer.
Unsupervised anomaly detection (USAD)~\cite{audibert2020usad} has two autoencoders consisting of one shared encoder and two independent decoders that train under a two-player training scheme: The autoencoder training phase and the adversarial training phase.
The LSTM-variational autoencoder (VAE)~\cite{park2018multimodal} employs a variational inference for reconstruction.
It performs by projecting multivariate observations and temporal dependencies at each time step into the latent space using an LSTM-based encoder, and decoding by estimating the expected distribution of multivariate inputs from the latent representation.
OmniAnomaly~\cite{su2019robust} applies VAE to model a time series signal as a probabilistic representation that predicts anomalies if the reconstruction likelihood of a given input is lower than the defined threshold.
The temporal hierarchical one-class network (THOC)~\cite{shen2020timeseries} consists of a multi-layer dilated recurrent neural network and a hierarchical deep support vector data description.
The graph deviation network (GDN)~\cite{deng2021graph} learns sensor relationships to detect deviations in anomalies from the learned pattern.
Regardless of temporal dependencies, it projects each sensor into its behavior embedding and considers the inter-correlation among sensors using graph attention (GAT) networks.
The multivariate time-series anomaly detection graph attention (MTAD-GAT) network~\cite{zhao2020multivariate} includes two parallel graph attention layers to learn the complex dependencies of multivariate time-series in both temporal and feature dimensions.
In addition, its approach jointly optimizes a forecasting-based model and a reconstruction-based model to obtain both short- and long-term representations.

It should be noted that none of these studies focus on extracting well-discriminated features in the latent space.
They merely infer a shared feature distribution to reconstruct the input or predict the observation on the next timestamp, thereby exhibiting potential problems such as reconstructing anomalies as they are or forecasting without recognizing long-term abnormal patterns.
\begin{figure*}[!ht]
    \centering
    \includegraphics[width=0.9\textwidth]{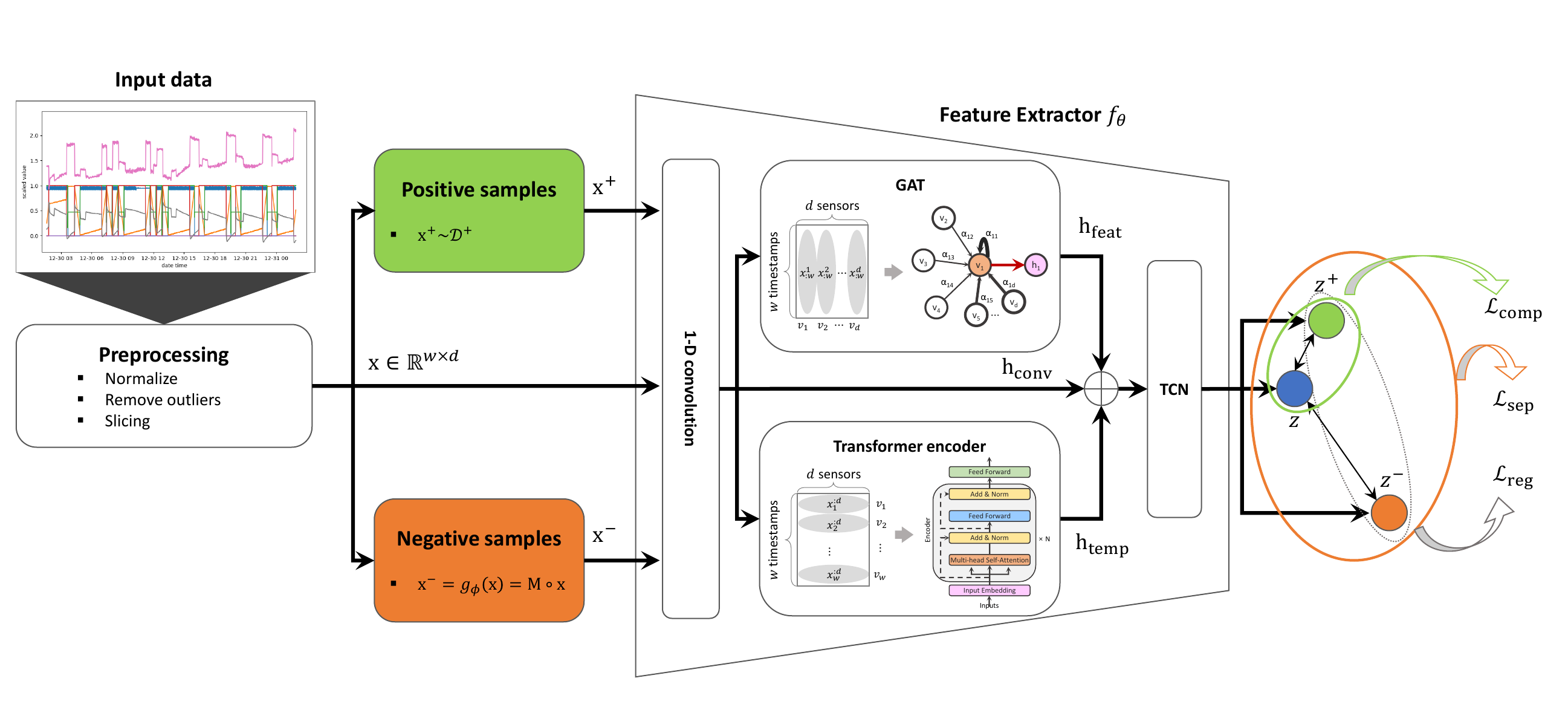}
    \caption{Overall architecture of LATAD.}
    \label{fig:architecture}
\end{figure*}
We were inspired by the recently proposed contrastive learning methods~\cite{chen2020simple, tonekaboni2021unsupervised} that utilize data augmentation to learn discriminative feature representations.
We considered that the feature extractor would distinguish between the normal and abnormal more effectively if we used more challenging and real-anomaly-like negative samples in its contrastive learning.
Therefore, we employed a learnable transformation that mines diverse and challenging negative samples to learn the normal representation in a more robust self-supervised manner for time-series anomaly detection.

\section{Proposed method}
\label{sec:method}
In this section, we first formalized the problem in subsection~\ref{subsec:problem}.
Subsection~\ref{subsec:latad} presents the formulation of the proposed method.
Subsection~\ref{subsec:training} describes the training procedure.
Finally, subsection~\ref{subsec:score} presents the proposed anomaly score from feature.

\subsection{Problem Formulation}
\label{subsec:problem}
A multivariate time series consists of multiple univariate time series from the same equipment or system (or broadly, an entity).
Anomaly detection in multivariate time series aims to detect anomalies at the entity level~\cite{su2019robust}.
The problem can be defined as follows.
An input of multivariate time-series is denoted by $\mathbf{x}\in\mathbb{R}^{d\times T}$, where $d$ and $T$ respectively represent the number of features in the input and the maximum length of timestamps.
We generated fixed-length inputs by a sliding window of length $W$ fragmenting a long time series.
The function of the time-series anomaly detection was to produce decisions $y\in\mathbb{R}^{T}$ across all timestamps, where $y_i\in\{0, 1\}$ indicates whether or not the $i$-th timestamp is an anomaly.

\subsection{LATAD}
\label{subsec:latad}
We addressed the abovementioned problem by extracting latent feature representations from input data via a feature extractor $f_\theta$ while simultaneously considering the inter-feature correlations and temporal dependencies in the sequential data.
In addition, we employed a triplet margin loss to learn the discriminative representations with positive samples from the temporal neighborhood and negative samples produced by a learnable generator.
Fig.~\ref{fig:architecture} illustrates the overall network architecture of LATAD.

\subsubsection{Preprocessing}
Given time series $X=\{\mathbf{x}_1,...,\mathbf{x}_T\}$, where $\mathbf{x}_t=\{x_t^1,...,x_t^d\}$ is an observation at each time step $t$ with $d$-variables.
These variables could be sensors in an industrial machine; we denoted them as \textit{features} for brevity.
There are five steps in the preprocessing stage for improving data efficiency.
First, replace empty or incorrect numeric data with the before and after values.
Second, down-sample the dataset as the time-series dataset is generally very long-term and requires a longer training time.
We averaged the values for $k$ steps as the signals were often received sparsely depending on the sensor.
Third, normalize the time series with the maximum and minimum values of the training data by considering that since the scale is different for each sensor, the model may incorrectly judge the importance:
\begin{equation}\label{eq:minmax}
    \mathbf{x}=\frac{\mathbf{x}_{\mathrm{raw}}-\mathrm{min}(X_{\mathrm{train}})}{\mathrm{max}(X_{\mathrm{train}})-\mathrm{min}(X_{\mathrm{train}})},
\end{equation}
where $\mathrm{max}(X_{\mathrm{train}})$ and $\mathrm{min}(X_{\mathrm{train}})$ represent the maximum and minimum values of the training set, respectively.
Fourth, since the model can respond sensitively to irregular and abnormal instances presented in the training data and consequently be trained in a biased manner, remove the outliers using the inter-quartile range (IQR) and linearly interpolate the removed points with the before and after values; IQR is the difference between the 3rd quarter (Q3) and 1st quarter (Q1) in the sorted data.
Finally, generate multiple training samples $\{\mathbf{x}_{1:w}\}$ that have been segmented using a sliding window.

\subsubsection{Extracting feature representations}
We used the following four modules to extract and fuse attributes with respect to inter-feature correlations and temporal dependencies: (in the order) 1-D convolution module, GAT module, transformer encoder module, and TCN module.
First, to extract the high-level features of each time-series input, we applied 1-D convolution with a kernel size of 5 in the first layer followed by the rectified linear unit ($\mathrm{ReLU}$) activation function.
Each position of the input data was multiplied by sliding the kernel and summing it, thereby resulting in the generation of feature maps with the size of the predefined output channel.
This process plays a role in extracting local features embedded in the data~\cite{zhang2015sensitivity}.
As a result, the module output was $\mathbf{h}_\mathrm{conv}\in\mathbb{R}^{w\times d}$.
The outputs of the 1-D convolution module were fed to the GAT module~\cite{velivckovic2017graph} and transformer encoder~\cite{vaswani2017attention} in parallel to model the feature relationships and measure the contribution between each observation on the time axis.

The GAT layer views the input data as a complete graph, where each vertex represents the values of one feature across all time stamps in the sliding window, and each edge corresponds the relationship between the two corresponding features.
Given a graph with $d$ vertices, $\{v_1, v_2,..., v_d\}$, where $v_i$ is the feature vector of each vertex,
the GAT layer computes the hidden representation for each vertex as follows:
\begin{equation}\label{eq:gat}
    h_i=\sigma(\sum_{j=1}^L\alpha_{ij}v_j),
\end{equation}
where $h_i$ denotes the hidden representation of vertex $i$, which has the same shape as input $v_i$; $\alpha_{ij}$ represents the attention score that measures the contribution of vertex $j$ to vertex $i$; $L$ denotes the number of adjacent vertices for vertex $i$; $\sigma$ denotes the sigmoid activation function, and the attention score $\alpha_{ij}$ is computed as follows:
\begin{equation}\label{eq:attention_e}
    e_{ij}=\mathrm{LeakyReLU}(w^\top\cdot(v_i\oplus v_j)),
\end{equation} 
\begin{equation}\label{eq:attention}
    \alpha_{ij}=\frac{\exp(e_{ij})}{\sum_{k=1}^{d}\exp(e_{ik})},
\end{equation}
where $\oplus$ denotes the concatenation of two vertex representations, $w\in\mathbb{R}^{2w}$ represents a column vector of learnable parameters, and $\mathrm{LeakyReLU}$ is a nonlinear activation function~\cite{xu2015empirical} that outputs $\max(0, x) + \text{slope}\times\min(0, x)$ for given input $x$.
Consequently, the GAT module output was $\mathbf{h}_{\mathrm{feat}}\in\mathbb{R}^{w\times d}$.


Contextual causality may exist within the sliding window, and in the case of a time series with short periodicity, one timestamp may contribute multiple non-adjacent timestamps.
The proposed method used a transformer encoder described in the original transformer work by~\citet{vaswani2017attention}; however, we did not use the decoder part of the architecture.
The encoder had two sublayers: the first was a multi-head self-attention mechanism and the second was point-wise fully connected (FC) network.
Residual connections existed around each of the two sublayers, followed by layer normalization.
Each preprocessed training sample $\mathbf{x}\in\mathbb{R}^{w\times d}$ constituted a sequence of $w$ feature vectors $\mathbf{x}_t\in\mathbb{R}^{d}$, \textit{i.e.} $\mathbf{x}\in\mathbb{R}^{w\times d}=\{\mathbf{x}_1,\mathbf{x}_2,...,\mathbf{x}_w\}$.
The original feature vector $\mathbf{x}_t$ was linearly projected onto a $d_{\mathrm{model}}$-dimensional vector space, the so-called embedding vector $\mathbf{u}_t\in\mathbb{R}^{d_{\mathrm{model}}}$ for $t\in\{1,...,w\}$, which were embedded concurrently by a single matrix-matrix multiplication; they formed the queries ($Q$), keys ($K$), and values ($V$) of the self-attention layer and were multiplied by the corresponding matrices as follows:
\begin{equation}\label{eq:selfattention}
    \mathrm{Attention}(Q, K, V)=\mathrm{softmax}(\frac{QK^T}{\sqrt{d_k}})V.
\end{equation}
Through the two-layer transformer encoders, the output was finally fed to the FC layer, which output $\mathbf{h}_\mathrm{temp}\in\mathbb{R}^{w\times d}$.

We concatenated the output hidden representations from the previous modules.
The concatenated representations, $\mathbf{h}_\mathrm{concat}=(\mathbf{h}_\mathrm{conv}\oplus\mathbf{h}_\mathrm{feat}\oplus\mathbf{h}_\mathrm{temp})\in\mathbb{R}^{w\times 3d}$, were processed by a TCN layer with $d_{\mathrm{model}}$ dimensions and four hierarchical layers to capture long-term sequential patterns and causalities in the time series.
Consequently, the reduced dimensional representation $\mathbf{z}\in\mathbb{R}^{d_{\mathrm{model}}}$ implied fused information from different modules, which were classified into anchor, positive, and negative feature representations according to the source, and then used as the ingredients for contrastive learning.

\begin{figure}[!t]
    \centering
    \subfloat[]{\includegraphics[width=0.5\linewidth]{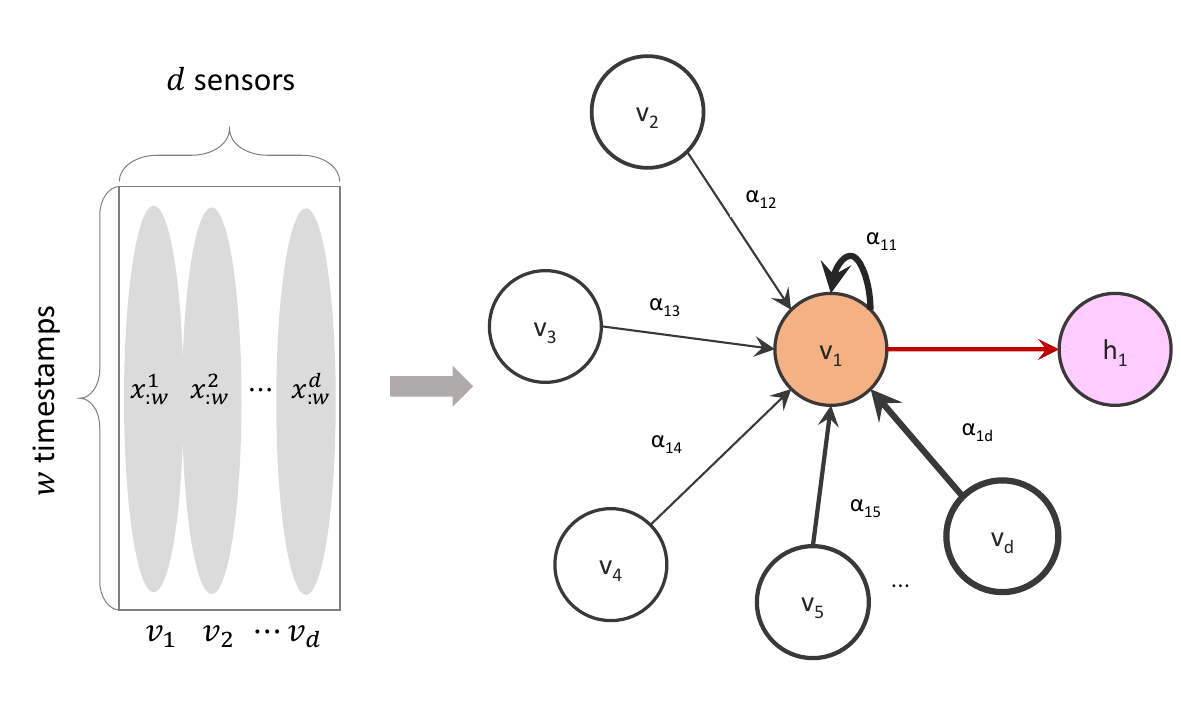}%
    \label{fig:gat}}
    \hfill
    \subfloat[]{\includegraphics[width=0.5\linewidth]{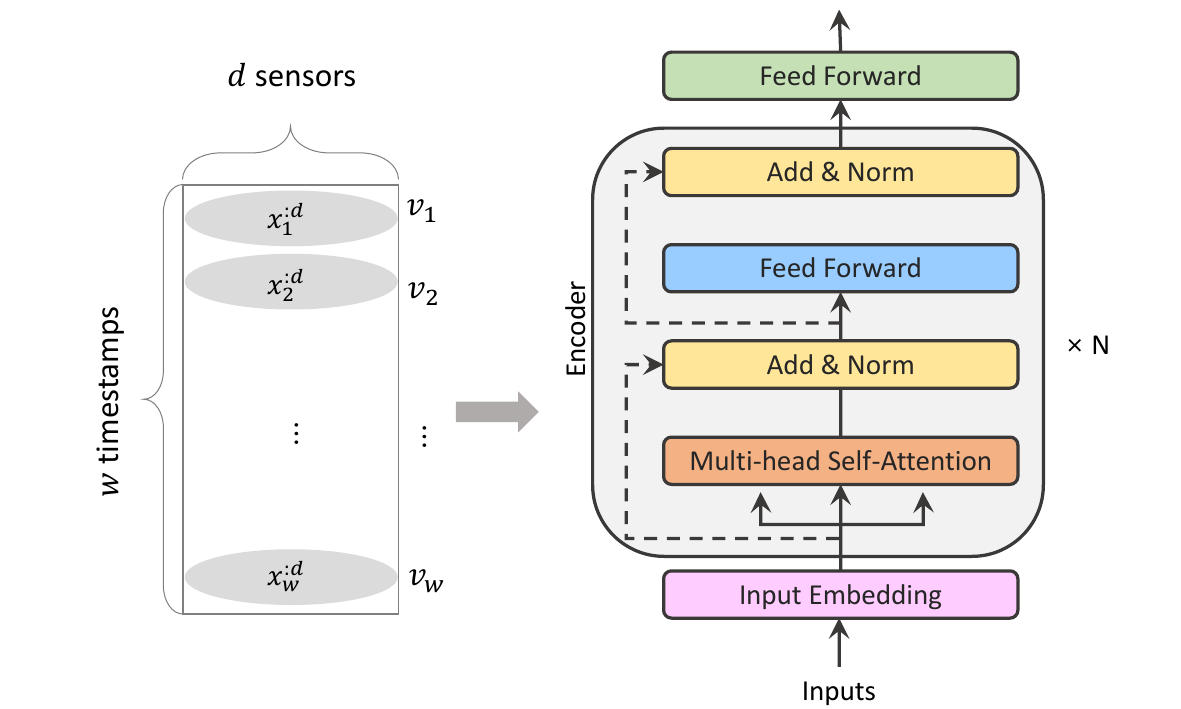}%
    \label{fig:transformerencoder}}
    \caption{Illustration of how the two attention-based modules to model correlations and temporal dependencies in the multivariate time-series data. (a) Graph attention module. (b) Transformer encoder module.}
\end{figure}

\subsubsection{Triplet-based contrastive learning}
We used triplet-based contrastive learning to implicitly estimate the data distributions using the positive and negative pairs on a reference known as the anchor (\textit{i.e.,} $\mathbf{z}=f_\theta(\mathbf{x})$).
The triplet objective ensured that similar time series exhibited similar latent features by minimizing the pairwise distance between the positive samples and maximizing it between the negative samples.

\begin{figure}[!t]
    \centering
    \subfloat[]{\includegraphics[width=0.5\linewidth]{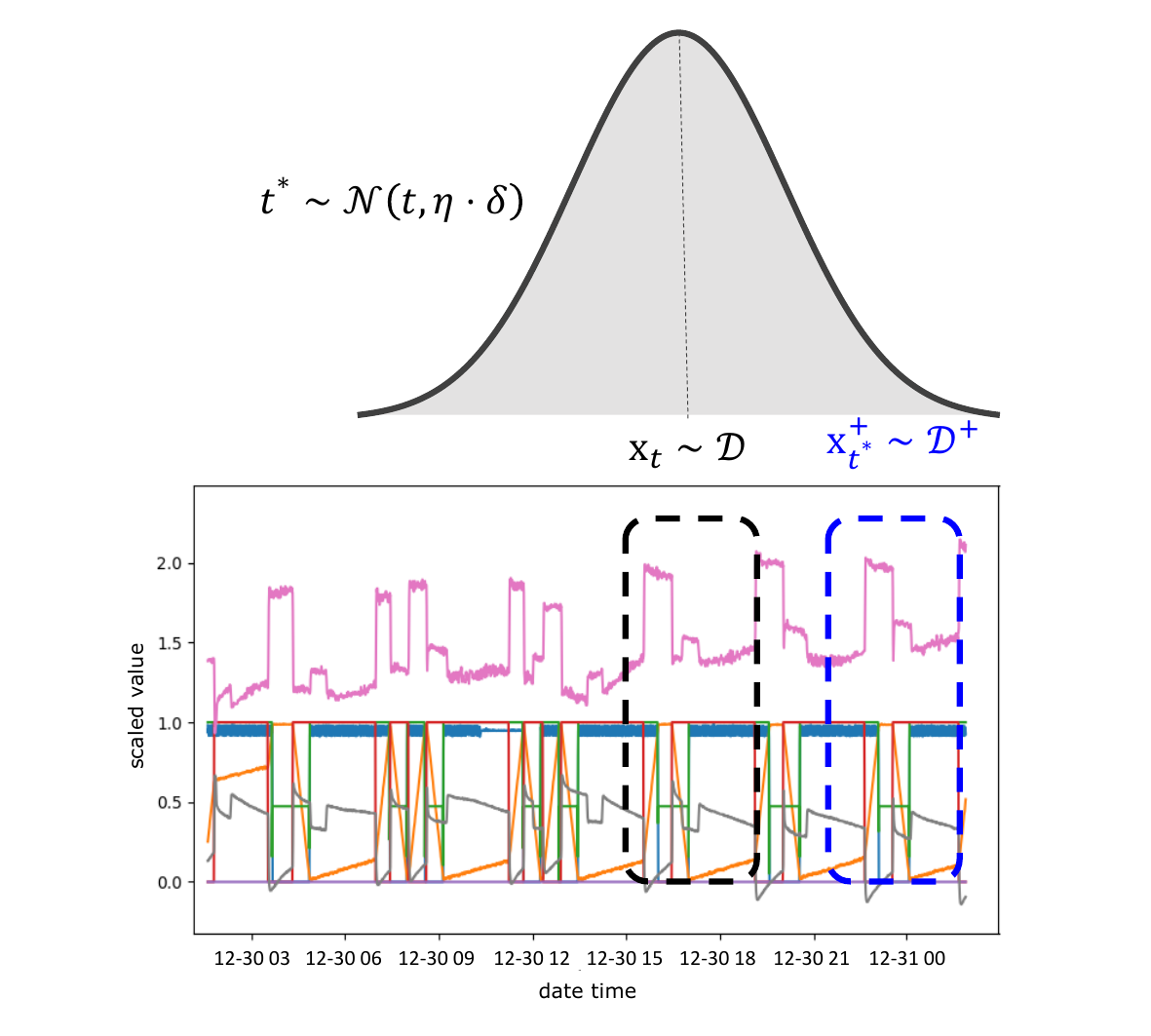}\label{fig:positivesamples}}%
    \hfill
    \subfloat[]{\includegraphics[width=0.5\linewidth]{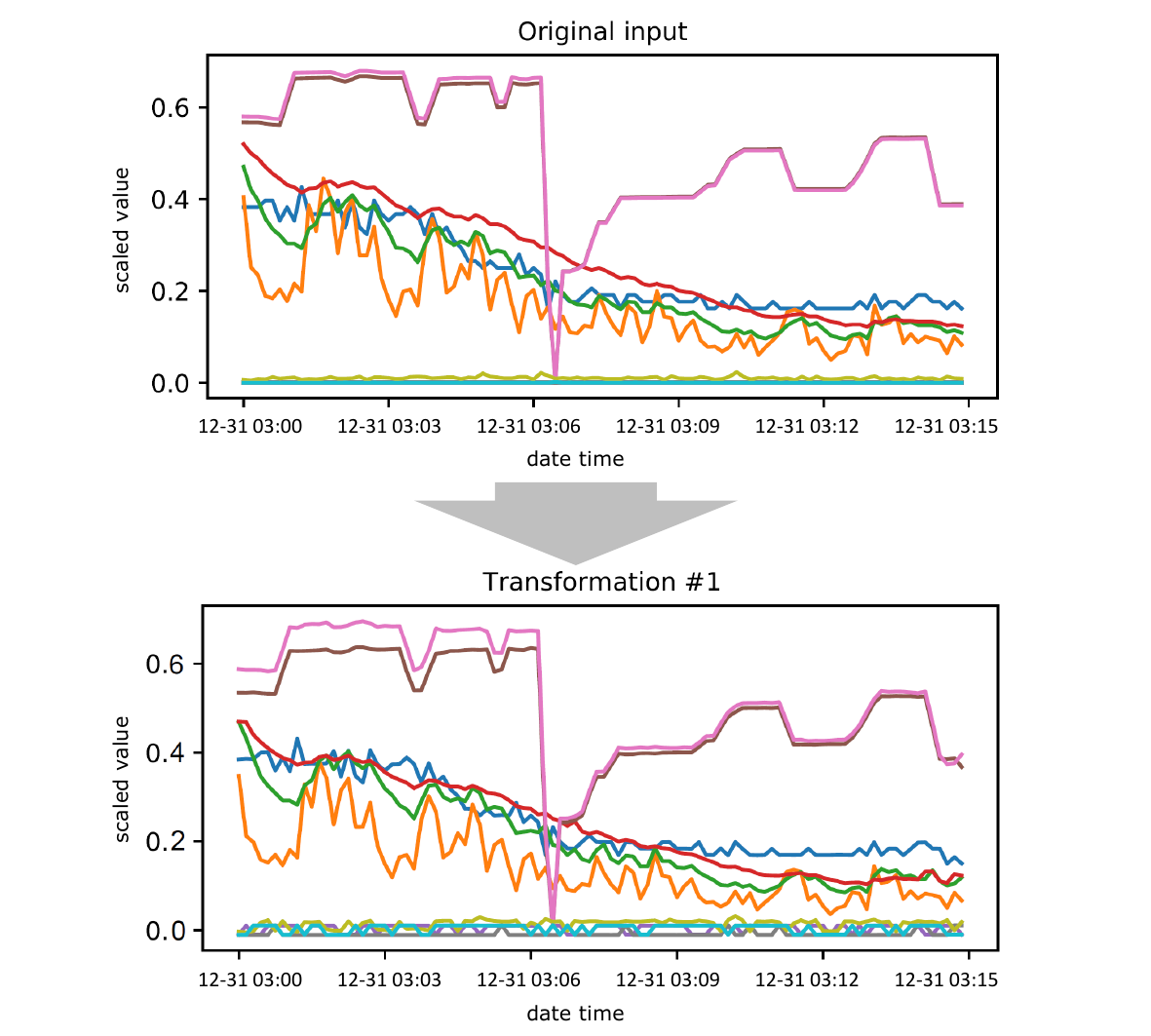}\label{fig:negativesamples}}%
    \caption{Triplet-based contrastive learning framework with positive samples and negative samples. (a) Positive samples from temporal neighborhood. (b) Generator-produced negative samples for the given input.}
    \label{fig:contrastive_latad}
\end{figure}

We utilized temporal neighborhoods for positive samples instead of randomly using several fixed transformations proposed in~\cite{chen2020simple,he2019momentum,grill2020bootstrap}, such as rotation, crop, and changing color, to take advantage of the contextual smoothness of the time series. 
Unlike images, neglecting the temporal context during augmentation may lose the semantic information in the time-series data with a respective correlation within two independent axes (time and feature).
Recently,~\citet{tonekaboni2021unsupervised} proposed representation learning of time-series data with augmentation; we followed their sampling method for acquiring positive samples.
They defined the temporal neighborhood of a reference time window $W_t$ as the set of all windows with centers $t$, sampled from a normal distribution $t^{*}\sim \mathcal{N}(t,\eta\cdot\delta)$, where $\mathcal{N}$ is a Gaussian centered at $t$, $\delta$ denotes the length of the sliding window, and $\eta$ represents the parameter that defines the range of the neighborhood.
The neighborhood boundaries $[t-\eta\cdot\frac{\delta}{2}, t+\eta\cdot\frac{\delta}{2}]$ were determined automatically using statistical testing, namely the augmented Dickey-Fuller (ADF) test.
We started from $\eta=1$ and iteratively increased the neighborhood size, while measuring the p-value from the ADF test at every step to find the neighborhood range around $W_t$.
If the p-value went above a certain threshold, it indicated that the signal was no longer stationary within this temporal region.
In this way, we found the widest neighborhood that was relatively stationary.
Consequently, several positive samples $\{\mathbf{x}^{+}_i\}_{i=1}^{N}$ were sampled from the temporal neighborhood, as illustrated in Fig.~\ref{fig:positivesamples}.
Hereinafter, we denote the latent feature representations of the positive samples as $\{\mathbf{z}_{i}^{+}\}_{i=1}^{N}=\{f_\theta(\mathbf{x}^{+}_i)\}_{i=1}^{N}$.

Since our problem statement was to detect anomalies in the time series, we considered an anomaly-like negative sample as a first choice.
Conventionally, time-series anomalies are classified into three main types: point, contextual, and collective.
For the given input data, fake-anomaly examples through several transformations such as cut-out, Gaussian noise, mix-up, adding trend, and temporal shift sufficiently reflect the three main types of time-series anomalies.
However, this cannot be guaranteed in a multivariate time series.
Multivariate time series requires additional consideration of the relationship between variables along the time axis. 
Increasing the number of variables causes more irregular abnormal patterns, thereby resulting in ambiguous to be classified.
Therefore, we employed learnable FC layers to transform the given input data $\mathbf{x}$.
The generator, denoted as $g_\phi(\cdot)$, jointly optimized with the feature extractor $f_\theta$ to effectively produce negative samples.
Assuming a preprocessed input sample $\mathbf{x}$, $N$ generators $\{g_{\phi_{i}}(\cdot)\}_{i=1}^{N}$ produce different masks $\{\mathrm{M}_{i}\}_{i=1}^{N}$ as follows:
\begin{equation}\label{eq:mask}
\begin{aligned}
    \mathrm{M}_{i}=\sigma(\mathrm{LeakyReLU}(\mathrm{w}_{i}^{(1)}\mathbf{x}+\mathrm{b}_{i}^{(1)})\mathrm{w}_{i}^{(2)}+\mathrm{b}_{i}^{(2)}),
\end{aligned}    
\end{equation}
where $\mathrm{w}_{i}^{(1)}$, $\mathrm{w}_{i}^{(2)}$, $\mathrm{b}_{i}^{(1)}$, and $\mathrm{b}_{i}^{(2)}$ are the learnable parameters in FC layers of the $i$-th generator for $i\in\{1,..,N\}$, and $\mathrm{LeakyReLU}$ and $\sigma$ represent the nonlinear activation functions.
As a result, the scale of $\{\mathrm{M}_{i}\}_{i=1}^{N}$ was set to 0 to 1.
We element-wise multiplied (denoted as $\circ$) with the input data to generate negative samples as follows:
\begin{equation}\label{eq:negative}
    \mathbf{x}_{i}^{-}=\mathrm{M}_{i}\circ\mathbf{x},\;\text{for}\;i\in\{1,..,N\}.
\end{equation}
They were fed to $f_\theta$ for extracting latent feature representations $\{\mathbf{z}_{i}^{-}\}_{i=1}^{N}={f_\theta(g_{\phi_{i}}(\mathbf{x}))}_{i=1}^{N}$.
The obtained negative samples based on learnable data augmentation were more diverse and productive than fixed transformations.



\subsection{Training Procedure}
\label{subsec:training}
We took advantage of the time divisions partitioned by the sliding windows as a supervisory signal to discriminate between normal and abnormal samples.
The proposed model was trained using the feature compactness loss, feature separateness loss, and the regularizer ($\mathcal{L}_{\mathrm{comp}}$, $\mathcal{L}_{\mathrm{sep}}$, and $\mathcal{L}_{\mathrm{reg}}$, respectively), and balanced by a hyperparameter $\lambda$ as follows:
\begin{equation}\label{eq:overallloss}
    \mathcal{L}_{\mathrm{LATAD}}=\mathcal{L}_{\mathrm{comp}}+\mathcal{L}_{\mathrm{sep}}+\lambda\mathcal{L}_{\mathrm{reg}}.
\end{equation}

\subsubsection{Feature compactness loss}
The feature compactness loss encourages positive samples to be close to the anchor to reduce intra-class variations.
It penalizes the discrepancies between them in terms of the cosine distance as:
\begin{equation}\label{eq:compact}
    \mathcal{L}_{\mathrm{comp}}=\mathbb{E}_{\substack{\mathbf{x}\sim\mathcal{D}\\ \mathbf{x}_{i}^{+}\sim\mathcal{D}^{+}}}\Bigg[\frac{1}{N}\sum_{i=1}^{N}\mathrm{dist}(\mathbf{z}, \mathbf{z}^{+}_{i})\Bigg],
\end{equation}
where $\mathrm{dist}(\mathbf{u}, \mathbf{v})$ is the adjusted cosine distance between the two vectors, which ranges from 0 to 1.
Since loss term makes all the items similar, all positive samples were mapped closely in the latent feature space, thereby losing the capability of preserving diverse patterns in normal data.
Nevertheless, adding temporal smoothness enhanced the ability of the model to minimize the intra-class variations, whereas jointly applying it with the separateness loss balanced its training.

\subsubsection{Feature separateness loss}
In the time-series domain, temporally close observations are likely to have similar distributions.
As we trained the model to extract the features, the feature compactness loss in (\ref{eq:compact}) draw positive samples and anchors close to each other.
In contrast, negative samples generated from the generators $\{g_{\phi_{i}}\}_{i=1}^{N}$ should be separated sufficiently far from the anchor in the latent feature space.
In addition, each negative sample should be different from one another to consider diverse patterns in the anomalies.
To address these challenges and simultaneously obtain shared feature representations between the anchor and positive samples, we proposed a feature separateness loss as follows:
\begin{equation}\label{eq:separate}
\begin{aligned}
    &\mathcal{L}_{\mathrm{sep}}=\\&\mathbb{E}_{\substack{\mathbf{x}\sim\mathcal{D}\\ \mathbf{x}_{i}^{+}\sim\mathcal{D}^{+}}}\Bigg[\frac{1}{N}\sum_{i=1}^{N}\max(0, \underbrace{\mathrm{dist}(\mathbf{z},\mathbf{z}^{+}_{i})}_\textrm{positive pair}
    -\underbrace{\mathrm{dist}(\mathbf{z}, \mathbf{z}^{-}_{i})}_\textrm{negative pair}+\epsilon_{i})\Bigg],
\end{aligned}
\end{equation}
where $\epsilon_{i}$ is $i$-th margin randomly selected from the range of 0.5 to 0.999 in the initialization phase.
Random margins allow the generator to produce diverse (semi) hard negative samples based on the assumption that anomalies exist in various ways, from easily identifiable to hard.
In addition, a small margin promotes initial learning, whereas a large margin later enhances subsequent learning.
Therefore, feature separateness loss enhances the discriminative capability of the proposed method.

\subsubsection{Regularizer}
Although we jointly optimized $f_{\theta}$ and $g_{\phi}$ using the separateness loss, as learning progressed, the generators produced negative samples that the model could easily distinguish positive and negative pairs, resulting in a decrease in discrimination.
To prevent this vulnerability, we constrained the generators so as to avoid the production of easy samples.
This was performed by employing the Kullback-Leibler divergence (KLD) as a regularizer to avoid the distribution of negative samples from falling too far from the distribution of positive samples in the feature space, as follows:
\begin{equation}\label{eq:kldivergence}
\begin{aligned}
    \mathcal{L}_{\mathrm{reg}}{}&=D_{KL}(\mathbf{z}^{+}||\mathbf{z}^{-})\\
    &=\mathbb{E}_{\substack{\mathbf{x}\sim\mathcal{D}\\ \mathbf{x}_{i}^{+}\sim\mathcal{D}^{+}}}\Bigg[\frac{1}{N}\sum_{i=1}^{N}\mathbf{z}^{+}_{i}\log \mathbf{z}^{+}_{i}-\mathbf{z}^{+}_{i}\log \mathbf{z}^{-}_{i}\Bigg].
\end{aligned}
\end{equation}

This regularizer affects the model performance but also the stability of the training process.
Like two-player adversarial training, $\mathcal{L}_{\mathrm{sep}}$ and $\mathcal{L}_{\mathrm{reg}}$ acts contrastive.
For training stability, we adjusted the influence of this regularizer with a weight parameter $\lambda$.
We observed that a small $\lambda$ made the model converge to sub-optimal too early, whereas a large one made the model training unstable.
We searched for a proper one and selected $\lambda=0.1$.

With the three objectives explained above, the pseudo-code of the training procedure of LATAD is provided in Algorithm~\ref{pseudo_code}.
\begin{algorithm}[!ht]
\SetAlgoLined
\SetKwInOut{Input}{Input}
\SetKwInOut{Output}{Output}
\Input{Unlabeled dataset $\mathcal{D}$, feature extractor $f_{\theta}$, generators $\{g_{\phi_{i}}\}_{i=1}^{N}$}
\Output{Feature extractor's parameters $\theta$}
 Preprocess input data $\mathbf{x}$\\
 Initialize $f_{\theta}$ and $\{g_{\phi_{i}}\}_{i=1}^{N}$\\
 Select margins $\{\epsilon_{i}\}_{i=1}^{N}$ randomly from [0.5, 0.999]\\
 \For{$e\in\mathrm{[1, max\_epoch]}$}{
  \For{$\mathrm{each\;mini\_batch}$ B}{
    $\mathbf{z}=f_{\theta}(\mathbf{x})$, where $\mathbf{x}\sim\mathcal{D}$\\
    $\mathbf{z}^{+}=\{f_{\theta}(\mathbf{x}_{i}^{+})\}_{i=1}^{N}$, where $\mathbf{x}_{i}^{+}\sim\mathcal{D}^{+}$\\
    $\mathbf{z}^{-}=\{f_{\theta}(g_{\phi_{i}}(\mathbf{x}))\}_{i=1}^{N}$\\
    Compute $\mathcal{L}_\mathrm{comp}$ on $(\mathbf{z}, \mathbf{z}^{+})$ by Equation~(\ref{eq:compact})\\
    Compute $\mathcal{L}_\mathrm{sep}$ on $(\mathbf{z}, \mathbf{z}^{+}, \mathbf{z}^{-})$ by
    Equation~(\ref{eq:separate})\\
    Compute $\mathcal{L}_\mathrm{reg}$ on $(\mathbf{z}^{+}, \mathbf{z}^{-})$ by Equation~(\ref{eq:kldivergence})\\
    Minimize (\ref{eq:overallloss}) with hyperparameter $\lambda$\\
    Update $\theta, \{\phi_{i}\}_{i=1}^{N}$ using ADAM~\cite{kingma2014adam}\\
  }
 }
 \caption{Training procedure for LATAD}
 \label{pseudo_code}
\end{algorithm}

\subsection{Anomaly Score}
\label{subsec:score}
LATAD estimates the extent of an anomaly by measuring its similarity in the feature space $\mathcal{Z}$, whereas standard reconstruction- or forecasting-based methods use an anomaly score defined in the input space $\mathcal{X}$.
We defined the score function for detecting anomalies, \textit{i.e.,} whether a given $\mathbf{x}$ is an anomaly or not, based on the feature representation $\mathbf{z}(\cdot)\in\mathbb{R}^{d_\mathrm{model}}$ learned by the proposed training objective.

First, we ran K-means clustering~\cite{macqueen1967some} on the features $\mathbf{z}(\mathbf{x}_m)$ using a cosine similarity, where $\mathbf{x}_m$ is a coreset (\textit{i.e.,} 10\%).
Thereafter, we measured the distance between the features for a given input and the K centers of the clusters.
The nearest distance was defined as the anomaly score $\mathcal{A}(\mathbf{x})$.
\begin{equation}\label{eq:score}
    \mathcal{A}(\mathbf{x})=\min_{m}\mathrm{dist}(\mathbf{z}(\mathbf{x}_m), \mathbf{z}(\mathbf{x}))/\|\mathbf{z}(\mathbf{x})\|.
\end{equation}
Intuitively, a low anomaly score indicates that $\mathbf {z}(\mathbf{x})$ is close to the representation of normal samples, whereas a higher anomaly score indicates that it is far from the representation of normal samples.
We calculated the anomaly scores and subsequently predicted, whether the time segments were normal and abnormal for all windows in the test dataset.
The ground truth binary label $y_t\in\{0,1\}$, which indicates whether a signal is an anomaly (1) or not (0), was provided only for the test dataset.
The predicted labels were obtained by comparing the anomaly scores $\mathcal{A}(\mathbf{x}_t)$ with a given threshold $\delta$ as follows:
\begin{equation}\label{eq:predict}
    \hat{y}_t=
    \begin{cases}
    1,\quad \text{if}\; \mathcal{A}(\mathbf{x}_t) > \delta\\
    0,\quad \text{otherwise},
    \end{cases}
\end{equation}
Although, several thresholding methods, such as the peaks over threshold, and dynamic threshold, have been adopted in previous works~\cite{su2019robust,hundman2018detecting} to compare the score, we utilized the brute-force method to make well-balanced decision between reducing false alarms and detecting all suspicious anomalies so that the proposed method searched through all possible thresholds that were over the average score of the validation set and selected the one that gave the highest performance.

\section{Experiments}
\label{sec:experiments}

\subsection{Datasets and Baselines}
We considered five benchmark real-world datasets in our experiments: Secure water treatement (SWaT)~\cite{mathur2016swat}, Water Distribution (WADI)~\cite{ahmed2017wadi}, Mars Science Laboratory (MSL)~\cite{hundman2018detecting}, Soil Moisture Active Passive (SMAP)~\cite{hundman2018detecting}, and server machine dataset (SMD)~\cite{su2019robust}, which are described below.
\begin{itemize}
    \item SWaT consists of multi-variate time-series data collected over 11 days from a scaled-down water treatment cyber-physical system. 
    The last 4 days of data contain 36 attacks.
    \item WADI consists of multi-variate time-series data of 16 days acquired from a scale-down city water distribution system with 123 sensors. 
    Out of 16 days, 14 days contain data under normal conditions, and 2 days under attack scenarios.
    \item MSL and SMAP are acquired from NASA spacecraft; they contain anomaly data from an incident surprise anomaly report for a spacecraft monitoring system. 
    \item SMD is made up of data from 28 server machines with 38 sensors for 10 days.
    The first 5 days consists of only normal data, and the last five days contains injected anomalies.
\end{itemize}
These datasets were effective in measuring the practicality of the model since they are all real-world data.
In addition, each dataset has different characteristics that helped measure the generalization ability of the model.
The statistics are presented in Table~\ref{tab:datasets}.
\begin{table}[!t]
    \centering
    \caption[Statistics of benchmark datasets]{The statistics of benchmark time-series anomaly detection datasets.}
    \begin{tabular}{c|rrrr}
         \toprule
         Dataset        &\multicolumn{1}{c}{Train}      &\multicolumn{1}{c}{Test}       &\multicolumn{1}{c}{Anomaly ratio}    &\multicolumn{1}{c}{\# Features}\\ 
         \midrule
         \textbf{SWaT}  &495,000    &449,919    &12.33\%        &51\\
         \textbf{WADI}  &784,537    &172,801    &5.77\%         &123\\
         \textbf{MSL}   &58,317     &73,729     &10.5\%         &55\\
         \textbf{SMAP}  &135,183    &427,617    &12.8\%         &25\\
         \textbf{SMD}   &25,300     &25,300     &4.21\%         &38\\
         \bottomrule
    \end{tabular}
    \label{tab:datasets}
\end{table}




\begin{table*}[!ht]
    \begin{center}
		\caption{\MakeUppercase{Anomaly detection accuracy in terms of \texttt{F1}, $\texttt{F1}_{\texttt{PA}}50$, and $\texttt{F1}_{\texttt{PA}}$.}}
    \resizebox{\linewidth}{!}{%
    \begin{tabular}{l|ccc|ccc|ccc|ccc|ccc}
    \toprule
    \multicolumn{1}{c}{\multirow{2}[4]{*}{Method}} & \multicolumn{3}{c}{SWaT} & \multicolumn{3}{c}{WADI} & \multicolumn{3}{c}{MSL} & \multicolumn{3}{c}{SMAP} & \multicolumn{3}{c}{SMD} \\
\cmidrule{2-16}          & \texttt{F1}    & $\texttt{F1}_{\texttt{PA}}50$ & $\texttt{F1}_{\texttt{PA}}$  & \texttt{F1}    & $\texttt{F1}_{\texttt{PA}}50$ & $\texttt{F1}_{\texttt{PA}}$  & \texttt{F1}    & $\texttt{F1}_{\texttt{PA}}50$ & $\texttt{F1}_{\texttt{PA}}$  & \texttt{F1}    & $\texttt{F1}_{\texttt{PA}}50$ & $\texttt{F1}_{\texttt{PA}}$  & \texttt{F1}    & $\texttt{F1}_{\texttt{PA}}50$ & $\texttt{F1}_{\texttt{PA}}$ \\
    \midrule
    DAGMM~\cite{zong2018deep} & 0.55 &   -    & 0.85 & 0.12 &    -   & 0.21 & 0.20 &   -    & 0.70 & \underline{0.33} &   -    & 0.71 & 0.24 &   -    & 0.72 \\
    LSTM-VAE~\cite{park2018multimodal} & 0.78 &   -    & 0.81 & 0.23 &  -     & 0.38 & 0.21 &    -   & 0.68 & 0.24 &    -   & 0.76 & 0.44 &   -    & 0.81 \\
    MSCRED~\cite{zhang2019deep} & 0.67 & 0.72 & 0.87 & 0.09 & 0.09 & 0.35 & 0.20 & 0.20 & 0.78 & 0.23 & 0.24 & 0.94 & 0.10 & 0.10 & 0.39 \\
    OmniAnomaly~\cite{su2019robust} & 0.78 & \underline{0.83} & 0.87 & 0.22 & 0.41 & 0.42 & 0.21 & 0.21 & 0.90 & 0.23 & 0.32 & 0.81  & 0.47 & 0.52 & 0.94 \\
    USAD~\cite{audibert2020usad}  & \underline{0.79} & 0.81 & 0.85 & 0.23 & 0.42 & 0.43 & 0.21 & 0.23 & 0.93 & 0.29 & \underline{0.32} & 0.82 & 0.43 & 0.51 & 0.94 \\
    THOC~\cite{shen2020timeseries}  & 0.61 & 0.68 & 0.88 & 0.13 & 0.18 & 0.51 & 0.16 & 0.16 & 0.89 & 0.12 & 0.12 & 0.78 & 0.17 & 0.21 & 0.54 \\
    GDN~\cite{deng2021graph}   & \textbf{0.81} & \textbf{0.84} & 0.94 & \textbf{0.57} & \textbf{0.71} & 0.86 & 0.22 & 0.23 & 0.90 & 0.25 & 0.25 & 0.71 & \underline{0.53} & \underline{0.57} & 0.72 \\
    MTAD-GAT~\cite{zhao2020multivariate} & 0.73 & 0.76 & 0.79 & 0.12 & 0.14 & 0.30 & \underline{0.28} & \underline{0.32} & 0.91 & 0.19 & 0.19 & 0.90 & 0.43 & 0.52 & 0.76 \\
    \midrule
    LATAD (ours) & 0.78 & 0.81 & 0.84 & \underline{0.40} & \underline{0.47} & 0.53 & \textbf{0.44} & \textbf{0.56} & 0.88 & \textbf{0.36} & \textbf{0.36} & 0.93 & \textbf{0.53} & \textbf{0.66} & 0.76 \\
    \bottomrule
    \end{tabular}
    }%
  \label{tab:performance}%
  \end{center}
\end{table*}%

We compared LATAD with the recently proposed representative time-series anomaly detection methods described in Section~\ref{sec:related}.
These include DAGMM~\cite{zong2018deep}, LSTM-VAE~\cite{park2018multimodal}, MSCRED~\cite{zhang2019deep}, OmniAnomaly~\cite{su2019robust}, USAD~\cite{audibert2020usad}, THOC~\cite{shen2020timeseries}, GDN~\cite{deng2021graph}, and MTAD-GAT~\cite{zhao2020multivariate}.

\subsection{Evaluation Metrics}
Several studies have reported the performance of time-series anomaly detection~\cite{zong2018deep, park2018multimodal, zhang2019deep, su2019robust, audibert2020usad, shen2020timeseries, deng2021graph, zhao2020multivariate,kim2021towards}.
We used the reported performance when available, or obtain them through reproduction.
For performance evaluation, three standard evaluation metrics were adopted: precision, recall, and F1-score.
They take the following form:
\begin{equation}\label{eq:metric}
\begin{aligned}
    \textrm{Precision} = \frac{\textrm{TP}}{\textrm{TP}+\textrm{FP}},
\end{aligned}
\end{equation}

\begin{equation}
\begin{aligned}
    \textrm{Recall} = \frac{\textrm{TP}}{\textrm{TP}+\textrm{FN}},
\end{aligned}
\end{equation}

\begin{equation}
\begin{aligned}
    \textrm{F1\text{-}score} = 2\cdot\frac{\textrm{Precision}\cdot \textrm{Recall}}{\textrm{Precision}+\textrm{Recall}},
\end{aligned}
\end{equation}
where TP, FP, and FN respectively denote the number of true positives, false positives, and false negatives.

Currently, most time-series anomaly detection methods measure the F1-score after applying a peculiar evaluation protocol named point adjustment (\texttt{PA}), proposed by~\citet{xu2018unsupervised}.
\texttt{PA} works as follows: if at least one moment in a contiguous anomaly segment is detected as an anomaly, the entire segment is considered to be correctly detected.
We denoted $\textit{\textbf{S}}$ as a set of $K$ anomaly segments: $\textit{\textbf{S}}=\{S_1, ..., S_K\}$, where $S_k=\{t_s^k,...,t_e^k\}$; $t_s^k$ and $t_e^k$ represent the start and end times, respectively.
The adjusted prediction labels were obtained as follows:
\begin{equation}\label{eq:adjustpredict}
    \hat{y}_t=
    \begin{cases}
    \multirow{2}{*}{1,} & \text{if} \; \mathcal{A}(\mathbf{x}_t) > \delta\\
                     & \text{or}\; t\in S_k \; \text{and} \mathop{\exists}\limits_{t' \in S_k}  \mathcal{A}(\mathbf{x}_{t'}) > \delta \\
    0, &\text{otherwise}.
    \end{cases}
\end{equation}
However, F1-score with an adjusted prediction (\textit{i.e.,}  $\texttt{F1}_{\texttt{PA}}$) may overestimate the model performance.
~\citet{kim2021towards} reported that even a randomly generated anomaly score drawn from a uniform distribution achieves the highest $\texttt{F1}_{\texttt{PA}}$ when compared to other state-of-the-art methods for most benchmark datasets; however, the point-wise F1-score is the poorest.
Therefore, we adopted $\texttt{F1}_{\texttt{PA}}k$, as well as original F1-score ($\texttt{F1}$) and adjusted F1-score ($\texttt{F1}_{\texttt{PA}}$), to rigorously evaluate the model performances.
$\texttt{F1}_{\texttt{PA}}k$ applied \texttt{PA} to $S_k$ only if the ratio of number of correctly detected anomalies in $S_k$ to its length exceeded $k$ (\%).

\subsection{Experimental Results}
For most scenarios, the proposed method exhibited comparable performance or outperformed the other baseline methods.
Table~\ref{tab:performance} lists the results obtained using the benchmark datasets. 
Overall, the later released models exhibited better performance, however, the results did not reveal no clear one-size-fits-all method for all datasets.

Except for the SWaT and WADI datasets, the proposed method updated the existing state-of-the-art methods.
In particular, for MSL, SMAP, and SMD datasets, the proposed method surpassed the second-highest performance by a substantial margin, thereby validating the effectiveness of its self-supervised learning discriminative features.
Averaging the results across all five datasets, LATAD exhibited a 6.53\% higher $\texttt{F1}$ score and a 13.17\% higher $\texttt{F1}_{\texttt{PA}}50$ score when compared to the best benchmark results.



\subsection{Anomaly Diagnosis}
We provided root cause identification to help users understand model decisions and control the defective systems.
We selected the top few univariate time series, ranked by their gradient scale as the root cause.
The premise was as follows: if a feature is salient, it is expected to have a significant impact on the model output when varied locally.
We can formulate the input gradients $\hat{\mathbf{g}}\in\mathbb{R}^{w\times d}$ for the given input data with the anomaly score~(\ref{eq:score}) as follows:
\begin{align}
\begin{split}
    \hat{\mathbf{g}}=&\nabla\mathcal{A}(\mathbf{x}){}=\frac{\partial\mathcal{A}(\mathbf{x})}{\partial\mathbf{x}}=\frac{\partial\mathcal{A}(\mathbf{x})}{\partial f_{\theta}(\mathbf{x})}\cdot\frac{{\partial f_{\theta}(\mathbf{x})}}{\partial\mathbf{x}}.
\end{split}
\end{align}\label{eq:gradient}
We performed a normalization of the gradient values of each sensor to prevent deviations arising from any one sensor being overly dominant over the other sensors, as follows:
\begin{equation}
    g'_i(t)=\frac{g_{i}(t)-\mu_{i}}{\sigma_{i}},
\end{equation}\label{eq:normalize}
where $\mu_i$ and $\sigma_i$ represent the mean and standard deviation across the timestamps of the $g_i(t)$ values respectively.
Therefore, the feature with the largest gradient magnitude (absolute value) was selected at each time step of the time axis to compute the overall abnormality at time step $t$, and the indices of those features were collected as candidates using the below given equation:
\begin{equation}
    R(t)=\argmax_{i}|g'_i(t)|.
\end{equation}\label{eq:candidate}
Finally, the top $k$ feature indices were selected based on their frequency of appearance from $\{R(t)\}_{t=1}^w$ and defined as the most likely root causes of an anomaly within the time window.

\begin{figure}
    \centering
    \subfloat[]{\includegraphics[width=0.45\linewidth]{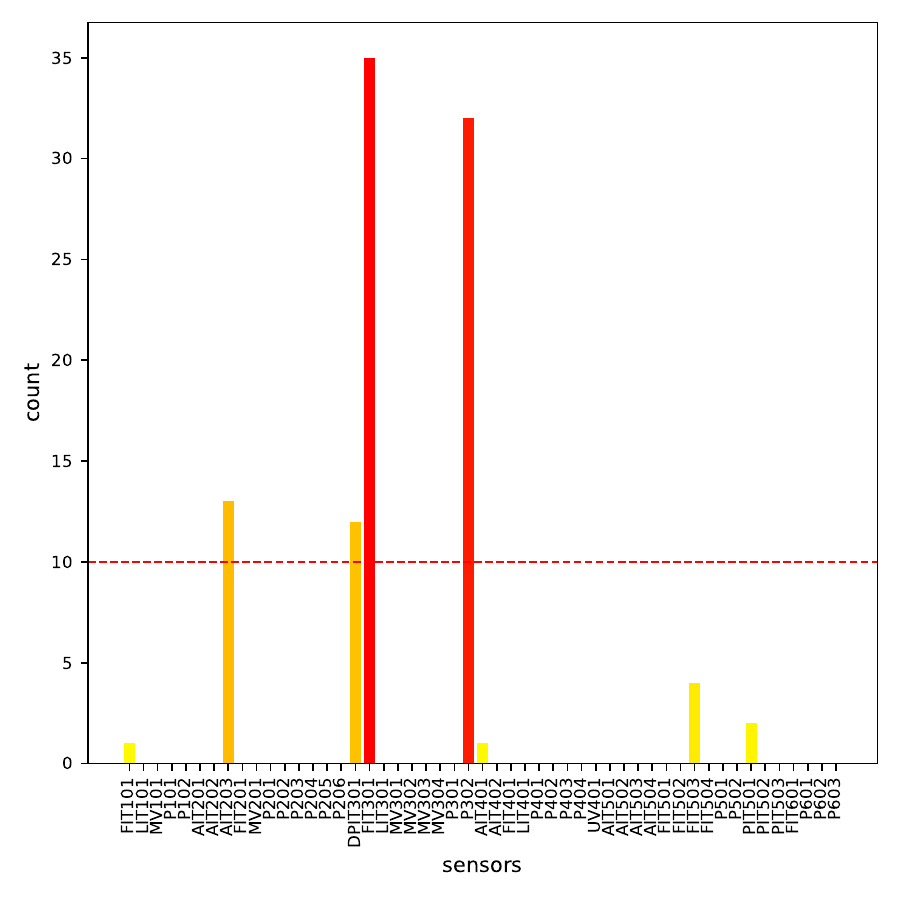}\label{fig:rootcauses}}
    \hfill
    \subfloat[]{\includegraphics[width=0.5\linewidth]{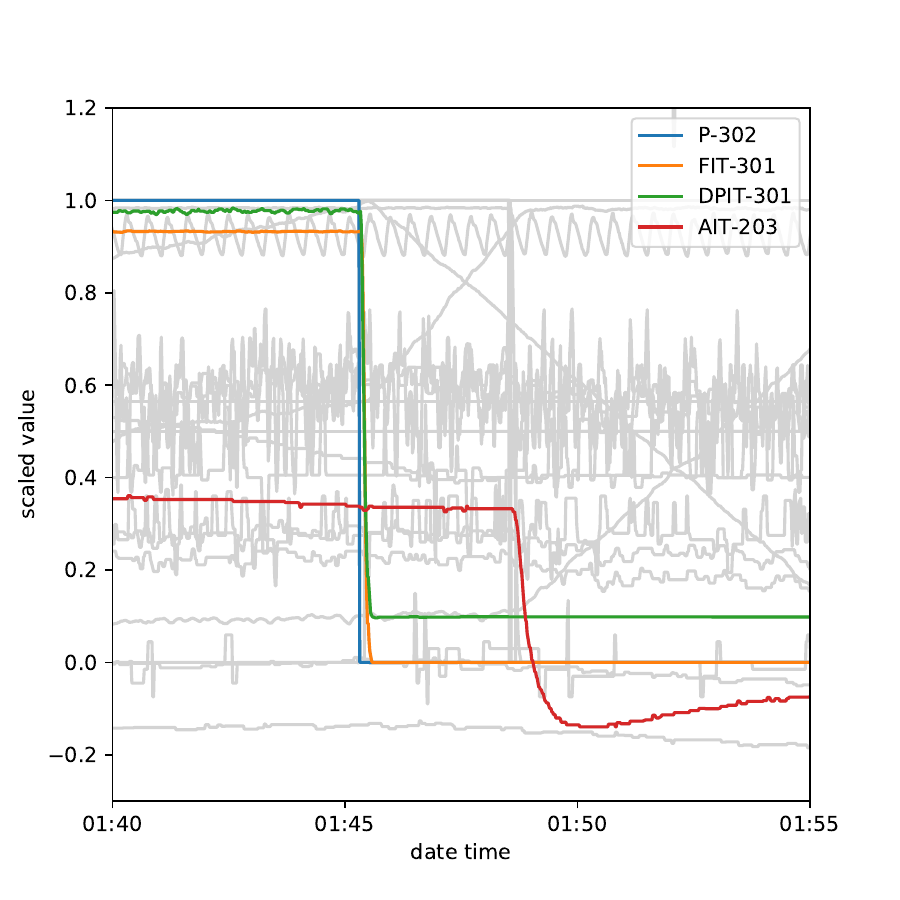}\label{fig:trendcharts}}
    \caption{Case study results of anomaly diagnosis. (a) Top $k$ root causes. (b) Trend charts of Top $k$ root causes on December 31, 2015, at 01:45:19.}
    \label{fig:top4}
\end{figure}

Fig.~\ref{fig:top4} shows the top four sensors identified based on the gradient as the root causes of failure that began on December 31, 2015, at 01:45:19 in the SWaT system.
This failure was carried out as part of a simulation training that blocks the flow of ultra-filtered water.
Within 5 minutes, selected sensors (\textit{i.e.} P-302, FIT-301, DPIT-301, and AIT-203) fluctuate sequentially.
Once the pump P-301 went down, the flow of purified water leading to the subsequent process is blocked, and the flow transmitter FIT-301 detected a decrease in flow rate.
Subsequently, the differential pressure indicator transmitter DPIT-301 detected differential pressure due to a decrease in flow rate.
Four minutes later, the oxidation-reduction potential of the chemical dropped in the previous process, and the analyzer indicator transmitter AIT-203 detected the abnormal concentration.

\section{Discussions}
\label{sec:discussions}
In this section, we describe an ablation study result to investivgate the effectiveness of our method and discuss on the relationship between the methods and datasets.

\subsection{Ablation Study}
We conducted a model evaluation by excluding each component $f_\theta$ to determine their individual contribution to the performance.
Table~\ref{tab:ablation_arthicture} represents the F1-score for the SWaT dataset when components are excluded one by one from the model.
As per the table, all the components contributed significantly to precision, \textit{i.e.,} the absence of even one module dropped the precision significantly.

Thereafter, we investigated how each loss affected the model performance by conducting an ablation study using $\mathcal{L}_{\mathrm{comp}}$ and $\mathcal{L}_{\mathrm{reg}}$.
Table~\ref{tab:ablation_loss} displays the corresponding results.
We observed that the exclusion of $\mathcal{L}_{\mathrm{comp}}$ from model training reduced the intra-class variance less and increased the normal range, resulting in a decrease in the recall.
Similarly, the generator produced only simple negative samples without $\mathcal{L}_{\mathrm{reg}}$.
These results implied that when the representation was loosely trained, model detected only certain anomalies and failed to identify ambiguous ones.
As a result, the precision increased whereas the recall decreased significantly.
\begin{table}[!t]
  \centering
  \caption{\MakeUppercase{Ablation study results: architecture.}}
    \begin{tabular}{lrrrr}
    \toprule
    \multicolumn{1}{c}{Architecture} & \multicolumn{1}{c}{Precison} & \multicolumn{1}{c}{Recall} & \multicolumn{1}{c}{F1} & \multicolumn{1}{l}{$\Delta\downarrow$ F1} \\
    \midrule
    LATAD (ours) & \textbf{0.9557} & \textbf{0.6586} & \textbf{0.7798} &  \\
    \hline
    w/o GAT & 0.8385 & 0.6893 & 0.7566 & -0.023  \\
    w/o TCN & 0.8318 & 0.6697 & 0.7355 & -0.044  \\
    w/o Transformer encoder & 0.8259 & 0.7035 & 0.7250 & -0.055  \\
    \bottomrule
    \end{tabular}%
  \label{tab:ablation_arthicture}%
\end{table}%

\begin{table}[!t]
  \centering
  \caption{\MakeUppercase{Ablation study results: loss.}}
    \begin{tabular}{lrrrr}
    \toprule
    \multicolumn{1}{c}{Architecture} & \multicolumn{1}{c}{Precison} & \multicolumn{1}{c}{Recall} & \multicolumn{1}{c}{F1} & \multicolumn{1}{l}{$\Delta\downarrow$ F1} \\
    \midrule
    LATAD (ours)\qquad\qquad\quad & \textbf{0.9557} & \textbf{0.6586} & \textbf{0.7798} &  \\
    \hline
    w/o $\mathcal{L}_{\mathrm{comp}}$   &0.9364   &0.6138 &0.7415 &-0.038\\
    w/o $\mathcal{L}_{\mathrm{reg}}$    &0.9719   &0.6064 &0.7468 &-0.033\\
    \bottomrule
    \end{tabular}%
  \label{tab:ablation_loss}%
\end{table}%

\subsection{Discriminative Feature Representations}
\begin{figure}[!h]
    \subfloat[]{\includegraphics[width=0.5\linewidth]{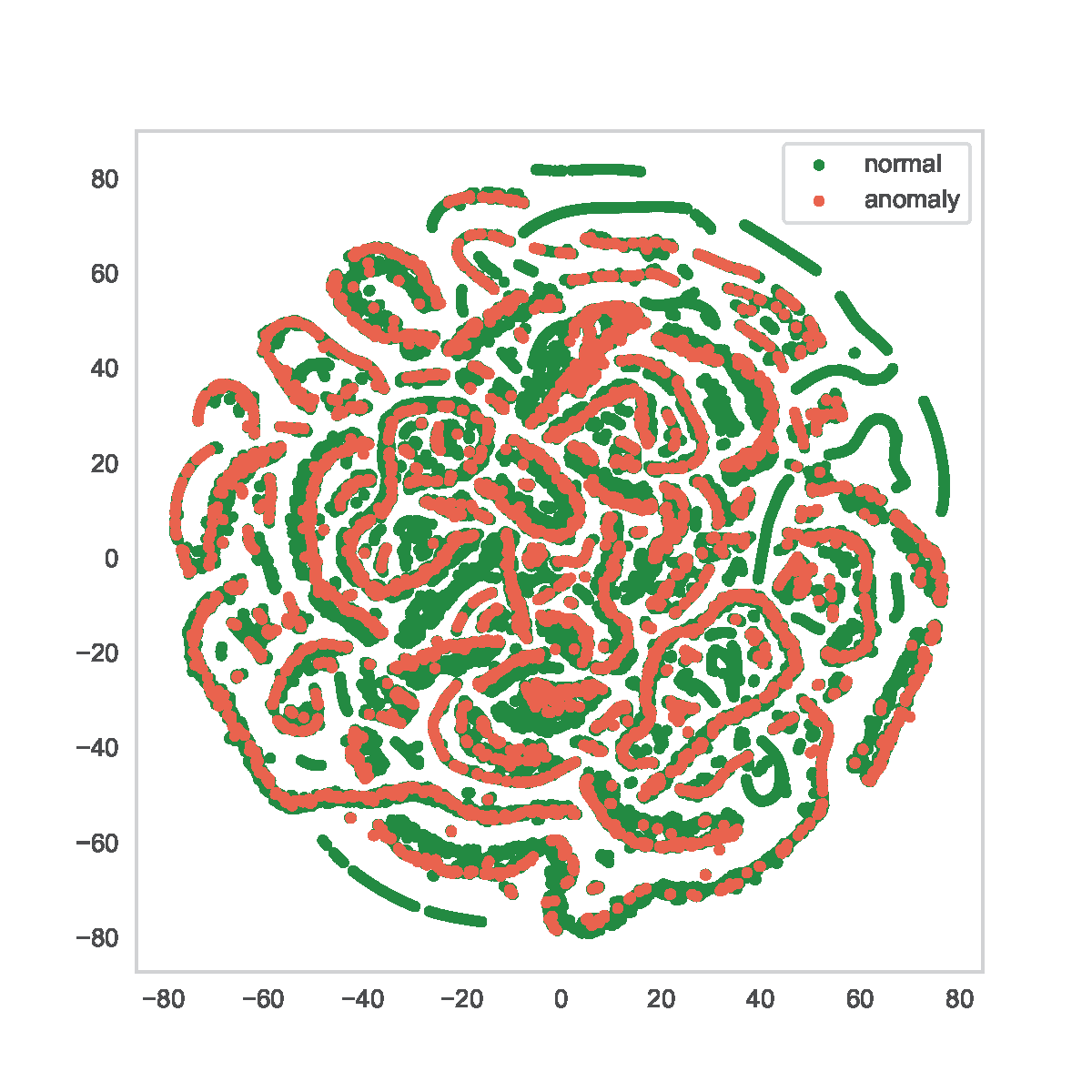}\label{fig:featuresMTADGAT}}
    \hfill
    \subfloat[]{\includegraphics[width=0.5\linewidth]{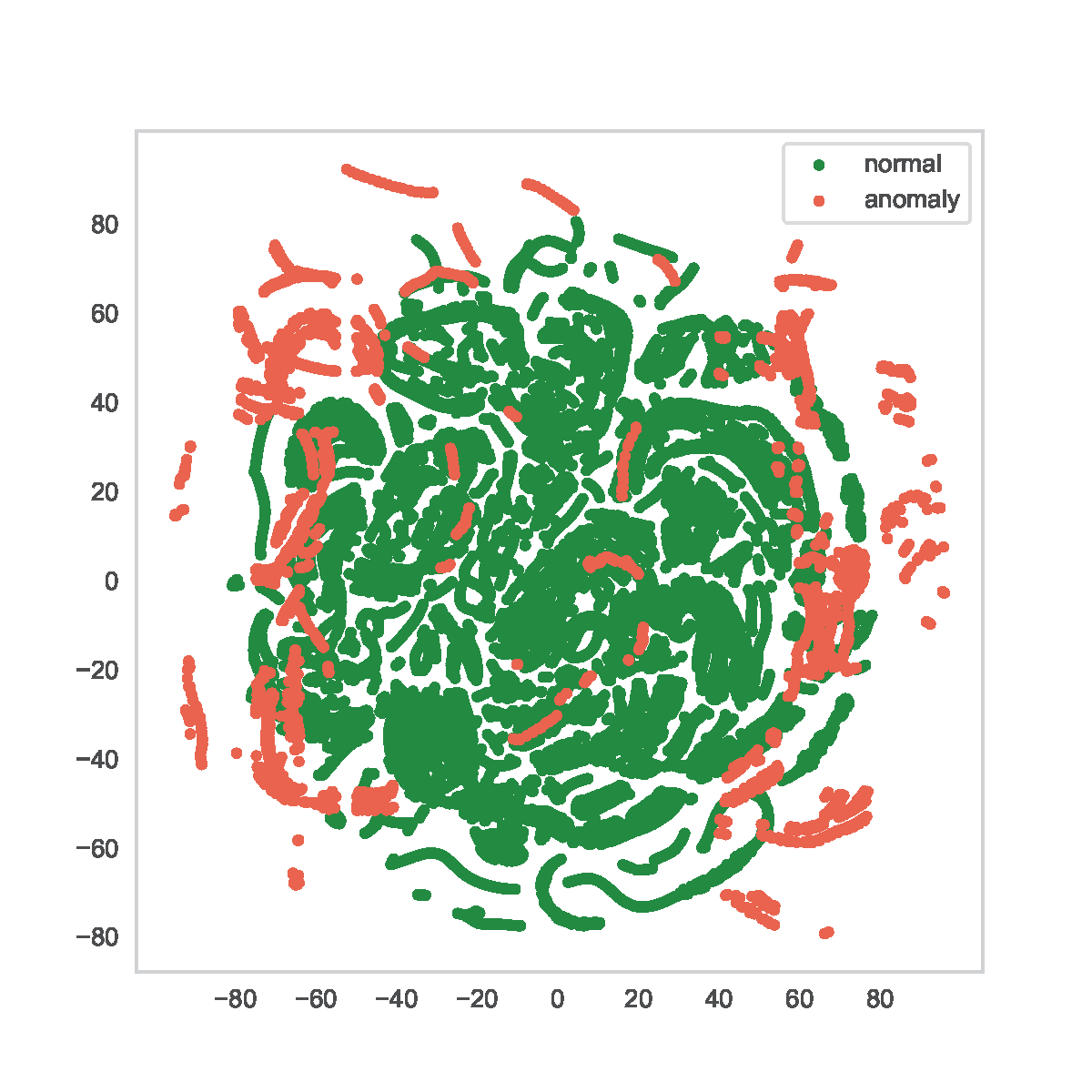}\label{fig:featuresLATAD}}
    \caption{Visualization of extracted feature representations for different methods (best viewed in color). (a) MTAD-GAT and (b) LATAD for the SWaT dataset. The green points are from normal class and red one are from anomalous class.}
	\label{fig:tsne}
\end{figure}

Fig.~\ref{fig:tsne} visualizes the learned latent feature representations of the SWaT test set using t-SNE~\cite{maaten2008visualizing} to compare the proposed method with MTAD-GAT~\cite{zhao2020multivariate} in terms of features discriminability. 
MTAD-GAT comprises architecture similar to the proposed method but employs forecasting and reconstruction objectives for anomaly detection.
The latent features that extracted by LATAD are clearly discriminated, whereas MTAD-GAT extracted shared features for reconstruction and forecasting; they are not well-discriminated in the latent feature space.

\subsection{Dependency on Data Characteristics}
Although the performance of the proposed model fell short for SWaT and WADI datasets, it outperformed the existing state-of-the-art methods for MSL, SMAP, and SMD.
Similarly, although GDN exhibited the best performance in SWaT and WADI datasets, it fell short of other models in the MSL and SMAP datasets.
Considering these results, we believed that there might be some association between the methods and datasets, and therefore attempted further explorations.

\begin{figure}[!t]
    \centering
    \subfloat[]{\includegraphics[width=0.47\linewidth]{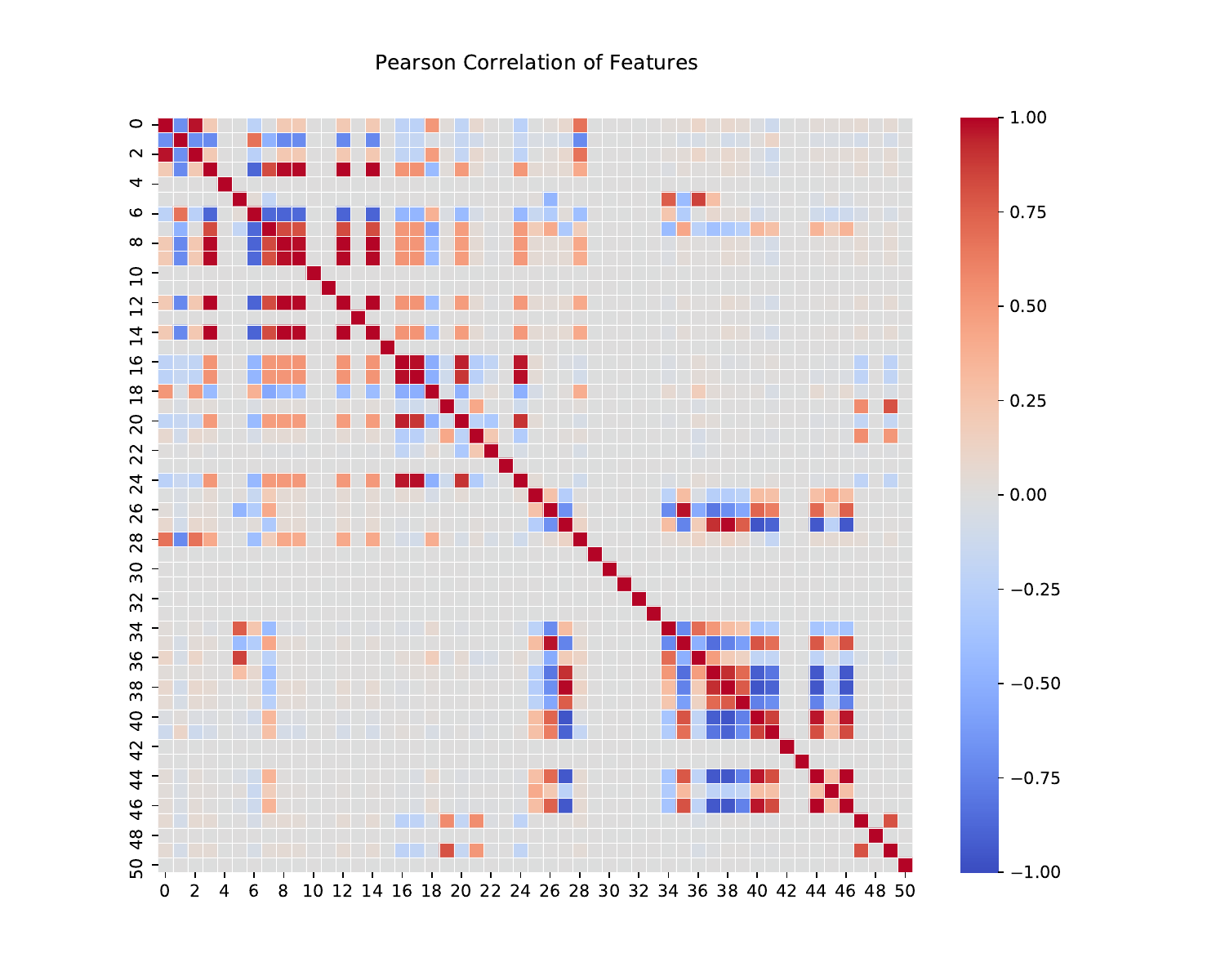}\label{fig:correlation_SWAT}}
    \hfill
    \subfloat[]{\includegraphics[width=0.47\linewidth]{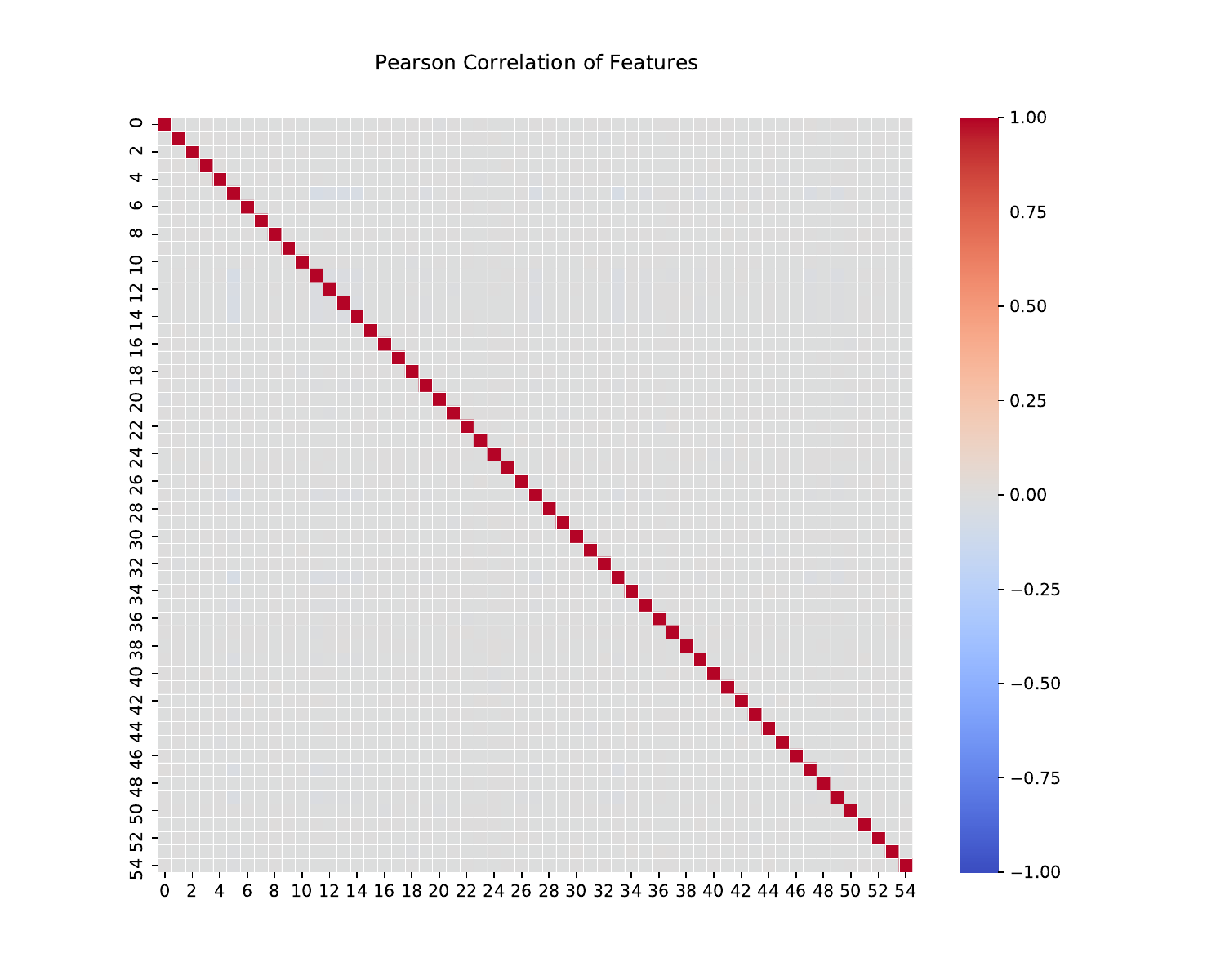}\label{fig:correlation_MSL}}
    \caption{Pearson correlation of features in (a) SWaT and (b) MSL.}
    \label{fig:correlation}
\end{figure}

First, we focused on the correlation between the features in the multivariate time-series data.
As per our observations, GDN captured inter-sensor relationships and detected anomalies that deviated from these relationships, thereby implying that GDN cannot extract meaningful information unless dataset features are highly correlated.
Fig.~\ref{fig:correlation} displays the Pearson correlation matrices of the features of the SWaT and MSL datasets; features of the SWaT dataset were highly correlated, wheareas those of MSL appeared to be independent.

\begin{figure}[!t]
    \centering
    \subfloat[]{\includegraphics[width=0.5\linewidth]{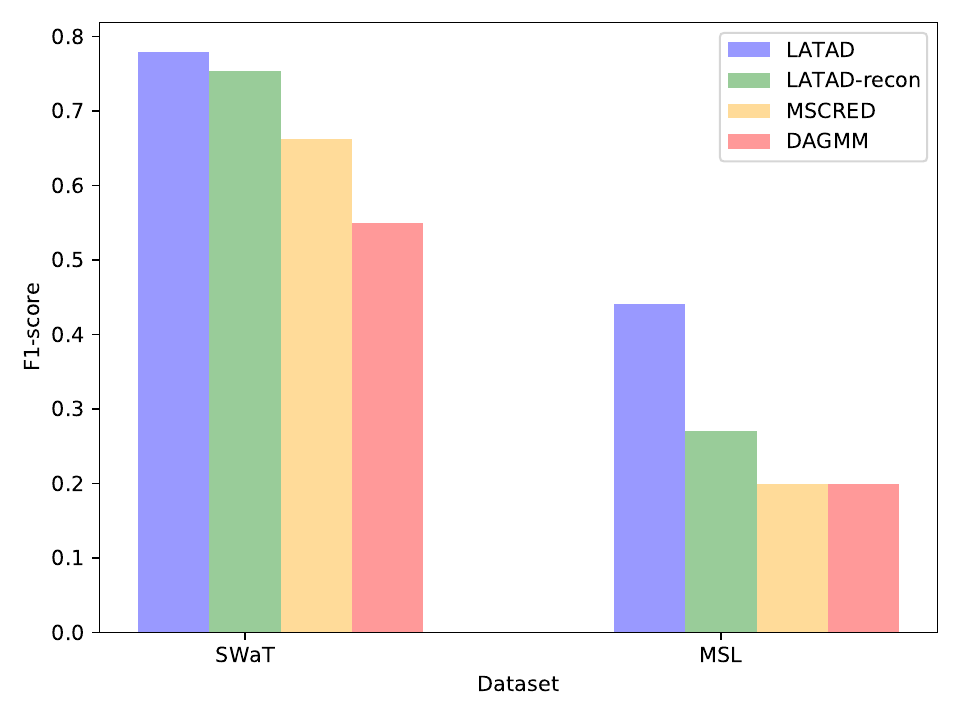}\label{fig:discuss_recon}}
    \hfill
    \subfloat[]{\includegraphics[width=0.5\linewidth]{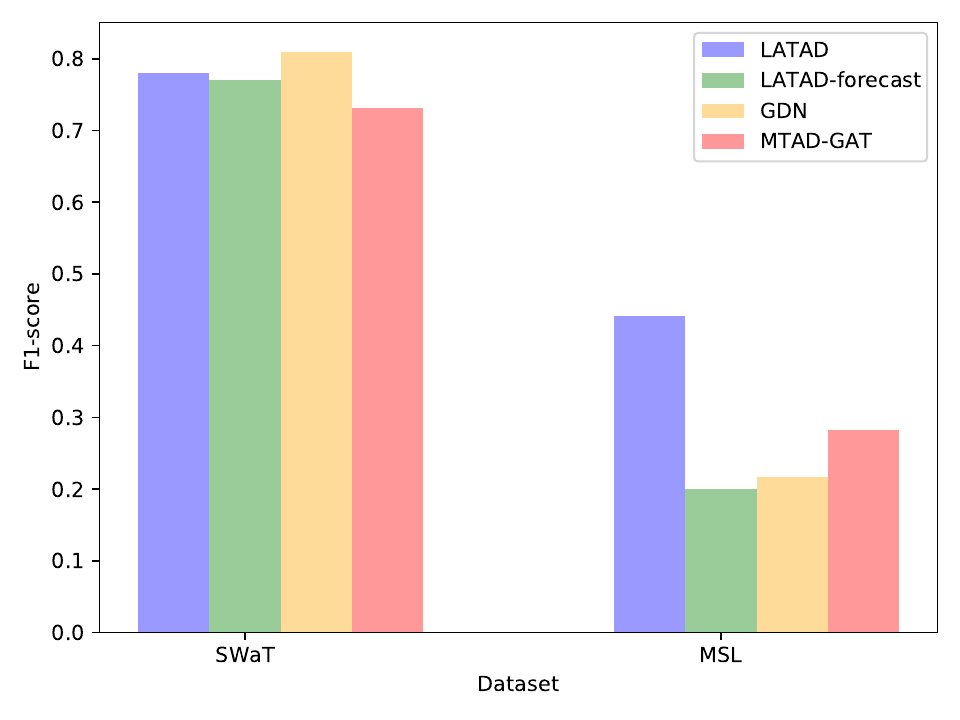}\label{fig:discuss_forecast}}
    \caption{Performance comparison between LATAD and (a) reconstruction- or (b) forecasting-based methods.}
    \label{fig:discussion}
\end{figure}

Second, we focused on the effects of contamination in the training dataset.
Unlike other datasets, MSL and SMAP contained unlabeled anomalies in the training data.
Reconstruction-based methods can be biased by the presence of anomalies in the training data, and may even reconstruct the actual anomalies of the test data.
In contrast, the proposed model extracts high-level discriminative features instead of generic features to reconstruct the input.
To prove this, we modified the proposed model by adding a decoder (\textit{i.e.,} LATAD-recon), and trained it to minimize the reconstruction error.
As shown in Fig.~\ref{fig:discuss_recon}, the corresponding model achieved the F1-score of 0.27 ($\downarrow$39\%).

Third, we tested the vulnerability of the proposed model for long-term contextual anomalies.
Since forecasting-based methods such as GDN and MTAD-GAT predict values that will likely come to the next step, they are good at catching rapidly deviating abnormal patterns but vulnerable to abnormal patterns that gradually change over a long period.
The MSL dataset comprises 47\% of contextual anomalies~\cite{hundman2018detecting}, and was therefore used for this experiment.
We modified the proposed model by adding an FC layer to forecast the next step (\textit{i.e.,} LATAD-forecast) and trained it to reduce the forecasting error.
Unlike SWaT, a clear degradation was observed in the F1-score ($\downarrow$55\%) compared to the original model for MSL (Fig.~\ref{fig:discuss_forecast}).

Based on these three examined aspects, we concluded that the proposed model exhibits generalization ability under various target system conditions.

\section{Conclusion}
In this study, we addressed the shortcomings of limited data conditions that affect the application of deep learning models in multiple industrial tasks by proposing a method for detecting time-series anomalies based on self-supervised learning.
The proposed learnable data-augment-based time-series anomaly detection (LATAD) robustly learned discriminative feature representations through a triplet-based contrastive learning that utilizes sampled positive examples from the temporal neighborhood and challenging negative samples produced by learnable neural networks.
This training scheme enhanced its generalization ability against data limitations.
In addition, LATAD employed a 1-D convolution layer, GAT, transformer encoder, and TCN to jointly consider the feature correlation and temporal dependencies.
Extensive assessments revealed the superior effectiveness and generalization ability of the proposed model when compared to previously established reconstruction- and forecasting-based methods.
Furthermore, we provided a gradient-based interpretation to help users localize anomalies and understand model decisions.

The proposed method advances time-series anomaly detection performance, but we think there is still room for improvement to apply to real-world scenarios as future work.
First, we can consider combining this approach with a forecasting-based strategy to land it on the real-time detection scenario.
The current sliding window-based method has an inevitable time delay for decision because it comprehensively assesses all data points in the window.
Second, we can consider online learning or continual learning strategy.
As time-series data is inherently non-stationary, we need to re-train the model periodically.
Most deep learning-based methods should provide a new batch of training data to fine-tune the model.
This training scheme may be problematic when the model should urgently adapt to the newly updated data distribution because it takes time to re-collect data after each data update.
\balance
\printbibliography

@article{chen2020simple,
  title={A simple framework for contrastive learning of visual representations},
  author={Chen, Ting and Kornblith, Simon and Norouzi, Mohammad and Hinton, Geoffrey},
  journal={arXiv preprint arXiv:2002.05709},
  year={2020}
}

@article{he2019momentum,
  title={Momentum contrast for unsupervised visual representation learning},
  author={He, Kaiming and Fan, Haoqi and Wu, Yuxin and Xie, Saining and Girshick, Ross},
  journal={arXiv preprint arXiv:1911.05722},
  year={2019}
}

@article{maaten2008visualizing,
  title={Visualizing data using t-SNE},
  author={Maaten, Laurens van der and Hinton, Geoffrey},
  journal={Journal of machine learning research},
  volume={9},
  number={Nov},
  pages={2579--2605},
  year={2008}
}

@inproceedings{zhang2019deep,
  title={A deep neural network for unsupervised anomaly detection and diagnosis in multivariate time series data},
  author={Zhang, Chuxu and Song, Dongjin and Chen, Yuncong and Feng, Xinyang and Lumezanu, Cristian and Cheng, Wei and Ni, Jingchao and Zong, Bo and Chen, Haifeng and Chawla, Nitesh V},
  booktitle={Proceedings of the AAAI Conference on Artificial Intelligence},
  volume={33},
  pages={1409--1416},
  year={2019}
}

@inproceedings{audibert2020usad,
  title={USAD: UnSupervised Anomaly Detection on Multivariate Time Series},
  author={Audibert, Julien and Michiardi, Pietro and Guyard, Fr{\'e}d{\'e}ric and Marti, S{\'e}bastien and Zuluaga, Maria A},
  booktitle={Proceedings of the 26th ACM SIGKDD International Conference on Knowledge Discovery \& Data Mining},
  pages={3395--3404},
  year={2020}
}

@article{park2018multimodal,
  title={A multimodal anomaly detector for robot-assisted feeding using an lstm-based variational autoencoder},
  author={Park, Daehyung and Hoshi, Yuuna and Kemp, Charles C},
  journal={IEEE Robotics and Automation Letters},
  volume={3},
  number={3},
  pages={1544--1551},
  year={2018},
  publisher={IEEE}
}

@inproceedings{su2019robust,
  title={Robust anomaly detection for multivariate time series through stochastic recurrent neural network},
  author={Su, Ya and Zhao, Youjian and Niu, Chenhao and Liu, Rong and Sun, Wei and Pei, Dan},
  booktitle={Proceedings of the 25th ACM SIGKDD International Conference on Knowledge Discovery \& Data Mining},
  pages={2828--2837},
  year={2019}
}

@inproceedings{zhao2020multivariate,
  title={Multivariate time-series anomaly detection via graph attention network},
  author={Zhao, Hang and Wang, Yujing and Duan, Juanyong and Huang, Congrui and Cao, Defu and Tong, Yunhai and Xu, Bixiong and Bai, Jing and Tong, Jie and Zhang, Qi},
  booktitle={2020 IEEE International Conference on Data Mining (ICDM)},
  pages={841--850},
  year={2020},
  organization={IEEE}
}

@article{pang2021deep,
  title={Deep learning for anomaly detection: A review},
  author={Pang, Guansong and Shen, Chunhua and Cao, Longbing and Hengel, Anton Van Den},
  journal={ACM Computing Surveys (CSUR)},
  volume={54},
  number={2},
  pages={1--38},
  year={2021},
  publisher={ACM New York, NY, USA}
}

@article{bai2018empirical,
  title={An empirical evaluation of generic convolutional and recurrent networks for sequence modeling},
  author={Bai, Shaojie and Kolter, J Zico and Koltun, Vladlen},
  journal={arXiv preprint arXiv:1803.01271},
  year={2018}
}

@article{bahdanau2014neural,
  title={Neural machine translation by jointly learning to align and translate},
  author={Bahdanau, Dzmitry and Cho, Kyunghyun and Bengio, Yoshua},
  journal={arXiv preprint arXiv:1409.0473},
  year={2014}
}

@inproceedings{vaswani2017attention,
  title={Attention is all you need},
  author={Vaswani, Ashish and Shazeer, Noam and Parmar, Niki and Uszkoreit, Jakob and Jones, Llion and Gomez, Aidan N and Kaiser, {\L}ukasz and Polosukhin, Illia},
  booktitle={Advances in neural information processing systems},
  pages={5998--6008},
  year={2017}
}

@inproceedings{woo2018cbam,
  title={Cbam: Convolutional block attention module},
  author={Woo, Sanghyun and Park, Jongchan and Lee, Joon-Young and Kweon, In So},
  booktitle={Proceedings of the European conference on computer vision (ECCV)},
  pages={3--19},
  year={2018}
}

@article{qin2017dual,
  title={A dual-stage attention-based recurrent neural network for time series prediction},
  author={Qin, Yao and Song, Dongjin and Chen, Haifeng and Cheng, Wei and Jiang, Guofei and Cottrell, Garrison},
  journal={arXiv preprint arXiv:1704.02971},
  year={2017}
}

@article{choi2021deep,
  title={Deep Learning for Anomaly Detection in Time-Series Data: Review, Analysis, and Guidelines},
  author={Choi, Kukjin and Yi, Jihun and Park, Changhwa and Yoon, Sungroh},
  journal={IEEE Access},
  year={2021},
  publisher={IEEE}
}

@inproceedings{hundman2018detecting,
  title={Detecting spacecraft anomalies using lstms and nonparametric dynamic thresholding},
  author={Hundman, Kyle and Constantinou, Valentino and Laporte, Christopher and Colwell, Ian and Soderstrom, Tom},
  booktitle={Proceedings of the 24th ACM SIGKDD international conference on knowledge discovery \& data mining},
  pages={387--395},
  year={2018}
}

@inproceedings{deng2021graph,
  title={Graph neural network-based anomaly detection in multivariate time series},
  author={Deng, Ailin and Hooi, Bryan},
  booktitle={Proceedings of the AAAI Conference on Artificial Intelligence},
  volume={35},
  number={5},
  pages={4027--4035},
  year={2021}
}

@article{tonekaboni2021unsupervised,
  title={Unsupervised representation learning for time series with temporal neighborhood coding},
  author={Tonekaboni, Sana and Eytan, Danny and Goldenberg, Anna},
  journal={arXiv preprint arXiv:2106.00750},
  year={2021}
}

@inproceedings{zong2018deep,
  title={Deep autoencoding gaussian mixture model for unsupervised anomaly detection},
  author={Zong, Bo and Song, Qi and Min, Martin Renqiang and Cheng, Wei and Lumezanu, Cristian and Cho, Daeki and Chen, Haifeng},
  booktitle={International conference on learning representations},
  year={2018}
}

@inproceedings{zaheer2020old,
  title={Old is gold: Redefining the adversarially learned one-class classifier training paradigm},
  author={Zaheer, Muhammad Zaigham and Lee, Jin-ha and Astrid, Marcella and Lee, Seung-Ik},
  booktitle={Proceedings of the IEEE/CVF Conference on Computer Vision and Pattern Recognition},
  pages={14183--14193},
  year={2020}
}

@article{zhang2015sensitivity,
  title={A sensitivity analysis of (and practitioners' guide to) convolutional neural networks for sentence classification},
  author={Zhang, Ye and Wallace, Byron},
  journal={arXiv preprint arXiv:1510.03820},
  year={2015}
}

@article{velivckovic2017graph,
  title={Graph attention networks},
  author={Veli{\v{c}}kovi{\'c}, Petar and Cucurull, Guillem and Casanova, Arantxa and Romero, Adriana and Lio, Pietro and Bengio, Yoshua},
  journal={arXiv preprint arXiv:1710.10903},
  year={2017}
}

@article{xu2015empirical,
  title={Empirical evaluation of rectified activations in convolutional network},
  author={Xu, Bing and Wang, Naiyan and Chen, Tianqi and Li, Mu},
  journal={arXiv preprint arXiv:1505.00853},
  year={2015}
}

@article{grill2020bootstrap,
  title={Bootstrap your own latent: A new approach to self-supervised learning},
  author={Grill, Jean-Bastien and Strub, Florian and Altch{\'e}, Florent and Tallec, Corentin and Richemond, Pierre H and Buchatskaya, Elena and Doersch, Carl and Pires, Bernardo Avila and Guo, Zhaohan Daniel and Azar, Mohammad Gheshlaghi and others},
  journal={arXiv preprint arXiv:2006.07733},
  year={2020}
}

@article{kingma2014adam,
  title={Adam: A method for stochastic optimization},
  author={Kingma, Diederik P and Ba, Jimmy},
  journal={arXiv preprint arXiv:1412.6980},
  year={2014}
}

@inproceedings{macqueen1967some,
  title={Some methods for classification and analysis of multivariate observations},
  author={MacQueen, James and others},
  booktitle={Proceedings of the fifth Berkeley symposium on mathematical statistics and probability},
  volume={1},
  number={14},
  pages={281--297},
  year={1967},
  organization={Oakland, CA, USA}
}

@inproceedings{mathur2016swat,
  title={SWaT: A water treatment testbed for research and training on ICS security},
  author={Mathur, Aditya P and Tippenhauer, Nils Ole},
  booktitle={2016 international workshop on cyber-physical systems for smart water networks (CySWater)},
  pages={31--36},
  year={2016},
  organization={IEEE}
}

@inproceedings{ahmed2017wadi,
  title={WADI: a water distribution testbed for research in the design of secure cyber physical systems},
  author={Ahmed, Chuadhry Mujeeb and Palleti, Venkata Reddy and Mathur, Aditya P},
  booktitle={Proceedings of the 3rd International Workshop on Cyber-Physical Systems for Smart Water Networks},
  pages={25--28},
  year={2017}
}

@article{shen2020timeseries,
  title={Timeseries anomaly detection using temporal hierarchical one-class network},
  author={Shen, Lifeng and Li, Zhuocong and Kwok, James},
  journal={Advances in Neural Information Processing Systems},
  volume={33},
  pages={13016--13026},
  year={2020}
}

@inproceedings{xu2018unsupervised,
  title={Unsupervised anomaly detection via variational auto-encoder for seasonal kpis in web applications},
  author={Xu, Haowen and Chen, Wenxiao and Zhao, Nengwen and Li, Zeyan and Bu, Jiahao and Li, Zhihan and Liu, Ying and Zhao, Youjian and Pei, Dan and Feng, Yang and others},
  booktitle={Proceedings of the 2018 World Wide Web Conference},
  pages={187--196},
  year={2018}
}

@article{kim2021towards,
  title={Towards a Rigorous Evaluation of Time-series Anomaly Detection},
  author={Kim, Siwon and Choi, Kukjin and Choi, Hyun-Soo and Lee, Byunghan and Yoon, Sungroh},
  journal={arXiv preprint arXiv:2109.05257},
  year={2021}
}

\begin{IEEEbiography}[{\includegraphics[width=1in,height=1.25in,clip,keepaspectratio]{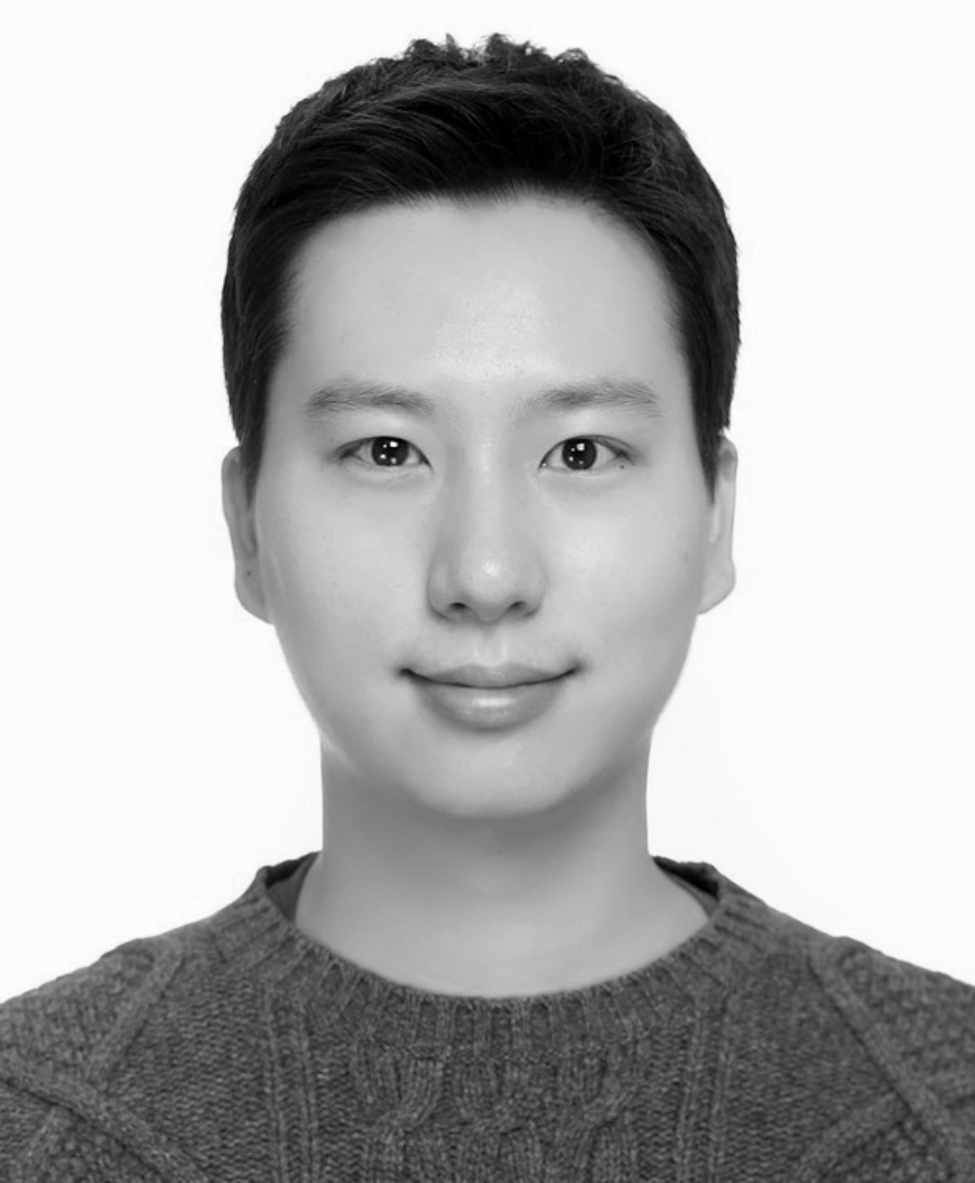}}]{Kukjin Choi}
received the B.S. degree in computer science and engineering from
Sogang University, Seoul, South Korea, in 2013, and the M.S. degree in electrical and computer engineering at Seoul National University, Seoul, South Korea, in 2022.
He is currently a staff software engineer and applied AI researcher in Innovation Center, Samsung Electronics.
His research interests include deep learning, anomaly detection, and time-series analysis.
\end{IEEEbiography}

\begin{IEEEbiography}[{\includegraphics[width=1in,height=1.25in,clip,keepaspectratio]{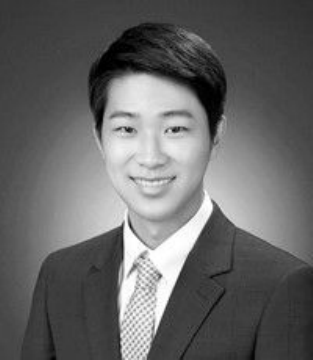}}]{Jihun Yi}
received the B.S. degree in electrical and computer engineering from Seoul National University, Seoul, South Korea, in 2017, where he is currently pursuing the Ph.D. degree in electrical and computer engineering. His research interests include deep learning, anomaly detection, and explaninable AI.
\end{IEEEbiography}

\begin{IEEEbiography}[{\includegraphics[width=1in,height=1.25in,keepaspectratio]{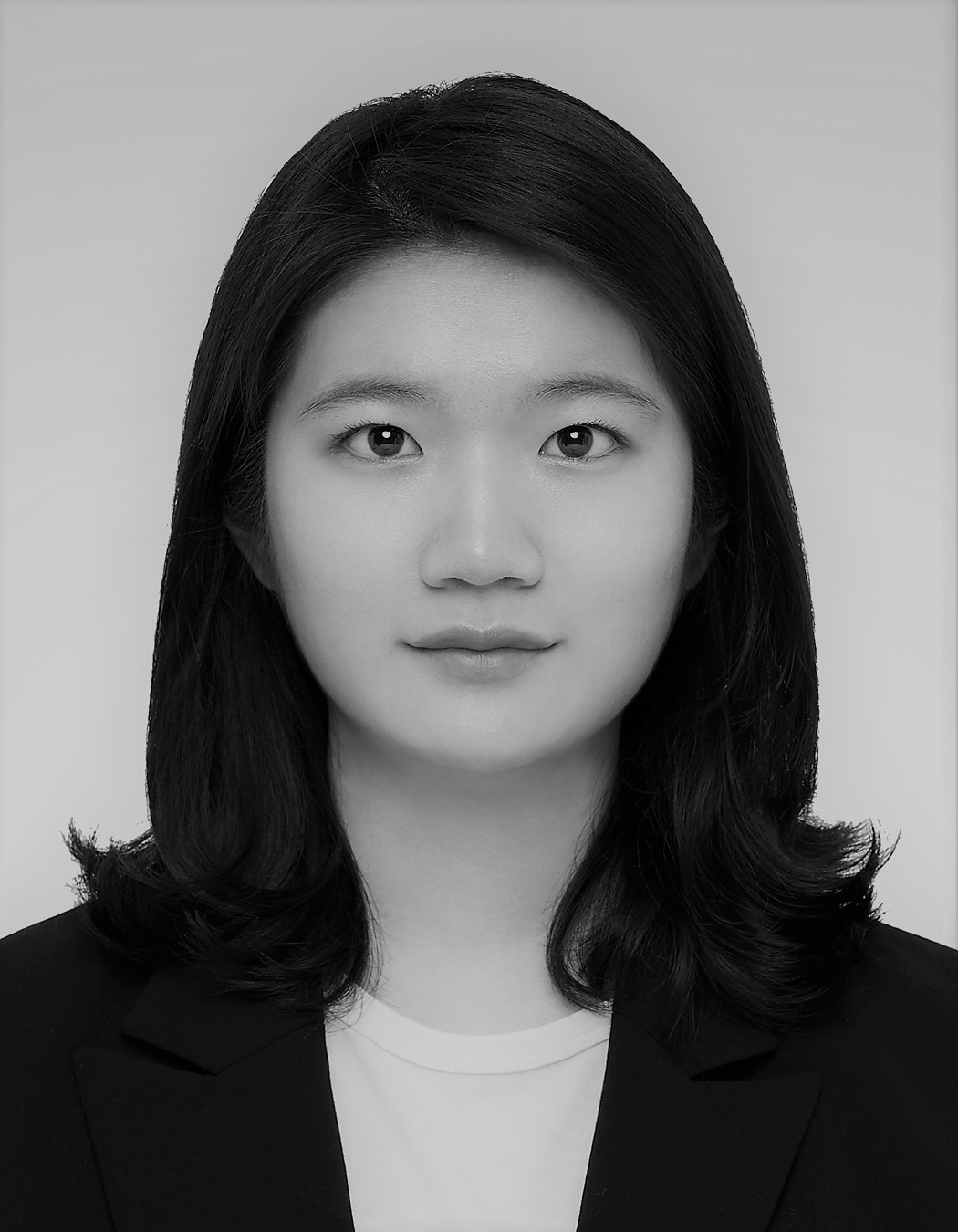}}]{Jisoo Mok}
received the B.S. degree in electrical engineering from California Institute of Technology, CA, USA, in 2018. She is currently pursuing the Integrated M.S./Ph.D. degree in electrical and computer engineering in Seoul National University, Seoul, South Korea. Her research interests include automated machine learning and uncertainty-aware deep learning.
\end{IEEEbiography}

\begin{IEEEbiography}[{\includegraphics[width=1in,height=1.25in,clip,keepaspectratio]{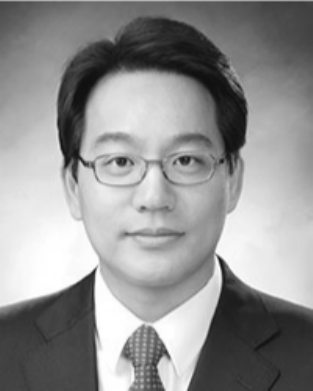}}]{Sungroh Yoon}
(S’99–M’06–SM’11)
received the B.S. degree in electrical engineering
from Seoul National University, South Korea,
in 1996, and the M.S. and Ph.D. degrees in
electrical engineering from Stanford University,
CA, USA, in 2002 and 2006, respectively. From
2016 to 2017, he was a Visiting Scholar with
the Department of Neurology and Neurological
Sciences, Stanford University. He held research
positions at Stanford University and Synopsys,
Inc., Mountain View, CA, USA. From 2006 to 2007, he was with Intel
Corporation, Santa Clara, CA, USA. He was an Assistant Professor with
the School of Electrical Engineering, Korea University, from 2007 to 2012.
He is currently a Professor with the Department of Electrical and Computer Engineering, Seoul National University. His current research interests
include machine learning and artificial intelligence. He was a recipient of
the SNU Education Award, in 2018, the IBM Faculty Award, in 2018,
the Korean Government Researcher of the Month Award, in 2018, the BRIC
Best Research of the Year, in 2018, the IMIA Best Paper Award, in 2017,
the Microsoft Collaborative Research Grant, in 2017, the SBS Foundation
Award, in 2016, the IEEE Young IT Engineer Award, in 2013, and many
other prestigious awards.
\end{IEEEbiography}

\vfill

\end{document}